\documentclass[pdflatex,sn-mathphys-num]{sn-jnl}

\usepackage{tabularray}

\usepackage{graphicx}%
\usepackage{multirow}%
\usepackage{amsmath,amssymb,amsfonts}%
\usepackage{amsthm}%
\usepackage{mathrsfs}%
\usepackage[title]{appendix}%
\usepackage{xcolor}%
\usepackage{textcomp}%
\usepackage{manyfoot}%
\usepackage{booktabs}%
\usepackage{algorithm}%
\usepackage{algorithmicx}%
\usepackage{algpseudocode}%
\usepackage{listings}%

\DeclareMathOperator*{\argmax}{arg\,max}

\begin{document}

\title[Extreme Weather Bench: A framework and benchmark for evaluation of high-impact weather]{Extreme Weather Bench: A framework and benchmark for evaluation of high-impact weather}


\author*[1,2]{\fnm{Amy} \sur{McGovern}}\email{dramymcgovern@gmail.com}

\author[2]{\fnm{Taylor} \sur{Mandelbaum}}

\author[1]{\fnm{Daniel} \sur{Rothenberg}}

\author[3]{\fnm{Nicholas} \sur{Loveday}}

\author[4]{\fnm{Corey} \sur{Potvin}}

\author[5]{\fnm{Montgomery} \sur{Flora}}

\author[6]{\fnm{Linus} \sur{Magnusson}}

\author[7]{\fnm{Eric} \sur{Gilleland}}

\author[8]{\fnm{John} \sur{Allen}}

\affil*[1]{\orgname{Brightband}, \orgaddress{\city{San Francisco}, \state{CA}, \country{USA}}}

\affil[2]{\orgdiv{School of Meteorology and School of Computer Science}, \orgname{University of Oklahoma}, \orgaddress{\city{Norman}, \postcode{73072}, \state{OK}, \country{USA}}}

\affil[3]{\orgname{Bureau of Meteorology}, \country{Australia}}

\affil[4]{\orgdiv{National Severe Storms Laboratory}, \orgname{National Oceanic and Atmospheric Administration}, \country{USA}}

\affil[5]{\orgname{The Weather Company}, \country{USA}}

\affil[6]{\orgname{European Centre for Medium Range Weather Forecasting},  \country{EU}}

\affil[7]{\orgdiv{Department of Statistics}, \orgname{Colorado State University}, \city{Fort Collins}, \state{CO}, \country{USA}}

\affil[8]{\orgdiv{ Earth and Atmospheric Sciences}, \orgname{Central Michigan University}, \city{Mount Pleasant}, \state{MI}, \country{USA}}

\abstract{Forecasting the wide variety of high-impact weather events experienced globally is a challenge for both Artificial Intelligence (AI) and Numerical Weather Prediction (NWP) models and it is critical that such models be properly verified before deployment. Although AI weather models are rapidly evolving (e.g.,\cite{GraphCast}, \cite{GenCast},\cite{FCN}, \cite{Pangu}), much of their evaluation is currently done either with a global-scale evaluation (e.g., \cite{WB},\cite{WB2}) or by hand-picking a small number of case studies or a region (e.g.,\cite{GenCast},\cite{Feldman2024},\cite{Liu2024}). A widely-used open-source benchmark suite focusing on high-impact weather will help to drive the science forward for all scales of weather models, as it has for other AI fields (e.g. \cite{ImageNet}). Here we introduce Extreme Weather Bench (EWB), a new community-driven benchmark suite that facilitates model validation and verification on a variety of high-impact hazards that matter to people around the globe. EWB provides a standard set of case studies (spanning across multiple spatial and temporal scales and different parts of the weather spectrum), observational data, impact-based metrics, and open-source code for users to evaluate their models. Verifying that a model works against a standard set of case studies, especially events that are high-impact for the general public, is a key piece of improving the trustworthiness of AI models \cite{McGovern2024}. EWB will help to drive the science forward for all weather models, enabling true comparisons across models and evaluating models on specific high-impact phenomena through the use of case studies. EWB is a free open-source community-driven system and will continue to evolve to include additional phenomena, test cases and metrics in collaboration with the worldwide weather and forecast verification community.}




\maketitle

\section{Introduction}\label{introduction}

Weather forecasting impacts every aspect of our lives, from everyday choices about clothing and whether to pack an umbrella, to life-saving decisions and actions in advance of high-impact weather. With the rapid proliferation of AI-based weather forecasts, it is increasingly important that we ensure these predictions are both ``trustworthy'' \cite{Cains2024, Wirz2024, Bostrom2024} and ``good'' \cite{Murphy1993}. However, defining what makes a forecast ``good'' is challenging and varies between individuals as well as by the phenomena being predicted.

Murphy \cite{Murphy1993} defined three categories in which we can measure the ``goodness'' of a forecast. First, a good forecast must be \emph{consistent}, which means that the forecast should correspond to the forecaster's best judgement of the forecast event. While Murphy focused on a human forecaster's judgment, the concept of ensuring an AI or NWP forecast is consistent over time is also a critical component of a trustworthy forecast \cite{Cains2024}. It is important that verification methods align with desirable behaviour of the forecasts. For probabilistic forecasts, evaluation using proper scoring rules \cite{Gneiting2007} can encourage consistent forecasts. For single-valued point forecasts, where a forecast directive is expressed in the form of a statistical functional such as, ``forecast the 90th percentile'', Gneiting \cite{Gneiting2011} formally defined the concept of ``consistent scoring functions''. For deterministic models, the notion of consistency is not formally defined, but can be encouraged by using verification methods that align with the core principles or goals of model development. Current evaluation systems of AI models often study only a handful of case studies, which does not address the consistency over time issue.

Murphy's second component of a good forecast was that it was of high \emph{quality}, meaning the forecast matched the observations of the actual event as closely as possible. This aspect is studied by most current evaluation and verification systems. There are a wide variety of metrics to measure different aspects of forecast quality \cite{Murphy1996, Brooks2024}. In most current AI evaluation systems, metrics such as the Root Mean Squared Error (RMSE) of key surface and model-level fields such as temperature and humidity or the Anomaly Correlation Coefficient (ACC) of 500mb geopotential heights take a prominent role, but a handful of studies are beginning to compare the phenomena generated by AI models to the known physics \cite{Hakim2024}.

Murphy's final quality of a good forecast is that it should provide high \emph{value}, which means it is of use to the end-users. This measure is often overlooked in favor of studying the quality of a forecast because many people assume that a high quality forecast automatically provides high-value. While they are correlated, it is not automatic that a forecast of high quality provides high value to the end-user, as it may not be predicting anything that impacts the user or it may not be focused on their specific needs. There is a growing body of research focused on working directly with forecasters to qualitatively understand what their needs are for forecasts \cite{Demuth2020, Ripberger2022} and a smaller body of research focusing on economic value of forecasts \cite{Richardson2000}.

As AI weather models propagate and are evaluated for possible operational use, a comprehensive and standardized ecosystem of model intercomparison, from global to local scales, is necessary. Forecast verification is an important topic that has been studied since at least 1884 \cite{Finley1884,Jolliffe2012}. As computer models and Numerical Weather Prediction (NWP) have provided additional value to forecasting, there have been a wide variety of different verification metrics and approaches developed. The World Meteorological Organization (WMO) has a joint working group, the WWRP/WGNE Joint Working Group on Forecast Verification Research (JWGFVR), focused on the issues of forecast verification worldwide\footnote{\url{https://community.wmo.int/en/activity-areas/wwrp/wwrp-working-groups/wwrp-forecast-verification-research}}.

The current state-of-the-art for global AI model evaluation primarily focuses on one or two scores, often Root Mean Squared Error (RMSE) or Anomaly Correlation Coefficient (ACC), and on prognostic variables. These metrics have several issues. First, they reward AI models for smooth fields, rather than fields that resemble physics-based weather. Second, they do not apply broadly across the high-impact weather spectrum. Third, they are often applied to variables at the 500mb height, rather than focused on the weather itself. ECMWF provides a standard evaluation framework for models\footnote{\href{https://www.ecmwf.int/en/forecasts/quality-our-forecasts}{Quality of our forecasts | ECMWF}} also used for their operational forecasts but it does not provide an in-depth evaluation of models across a variety of scales. Additionally, it focuses on broad metrics and not on any specific test cases. WeatherBench \cite{WB,WB2} is quickly becoming a standard for evaluation at the large-scale but it also lacks case studies, and evaluation across a variety of high-impact hazards. \cite{Olivetti2024} and 70 focused on specific extremes using a WeatherBench style approach but did not look at specific events or use impact-based metrics. In addition, there are several efforts focused on organizing metrics by model type and use-case, including a Unified Forecasting Community workshop\footnote{\href{https://dtcenter.org/events/2021/2021-dtc-ufs-evaluation-metrics-workshop/final-metrics-lists}{2021 DTC UFS Evaluation Metrics Workshop - Final Metrics Lists}}, the World Meteorological Organization (WMO) WWRP/WGNE JWGFVR, the WMO WP-MIP: the Weather Prediction Model Intercomparison Project\footnote{\href{https://www.wcrp-esmo.org/activities/wp-mip}{WP-MIP: the Weather Prediction Model Intercomparison Project --- ESMO Website}}, and the \texttt{scores} package \cite{scores_package}.

This work presents Extreme Weather Bench (EWB), the first community-driven benchmark suite designed to verify consistency and quality of a forecast for both AI and traditional NWP models, focusing on high-impact extreme events. EWB provides a timely and novel benchmark suite and approach to forecast verification focused on high-impact weather phenomena and verification based on observations, whenever possible. We will be adding additional phenomena, data, test cases and metrics in additional collaborations with the worldwide weather and forecast verification community, as well as building an extension for evaluating probabilistic forecasts. Through this worldwide involvement, we aim for EWB to become a standard testing suite across AI models. Further, the development of EWB, both in terms of data availability and code availability, prioritized open-source access and ease of use. Understanding the ability of models to predict extreme weather and climate hazards cannot be limited to select domain experts, given the material impact such events have on life and property. To that end, EWB also provides a benchmark suite that can spur scientific innovation in AI for weather, something that has been seen across other fields including ImageNet \cite{ImageNet} for AI and vision and the recent FHIBE \cite{Xiang2025}, Critical Assessment of Structure Prediction \cite{CASP} for protein folding, and General Language Understanding Evaluation \cite{GLUE} for language models.

\section{Extreme Weather Bench}\label{extreme-weather-bench}

EWB's high-impact phenomena span multiple spatial and temporal scales and different parts of the weather spectrum, ranging from short-term and small-scale impacts such as severe storms to long-term and larger-scale impacts such as heat waves. Focusing on global impact, we provide case studies and data for events around the world. By providing a wide variety of case studies across many high-impact phenomena, EWB can help evaluate the \emph{consistency} and \emph{quality} of the AI forecasts and also address the grey swan challenge \cite{LinGreySwan2016} and the questions of whether AI can effectively forecast extremes \cite{Nath2026}. Evaluating only on case studies can lead to the ``forecaster's dilemma'' \cite{Lerch2017}, where a system may overforecast to achieve high scores on the impactful events but then perform poorly overall. EWB mitigates this for many of the events by also providing data for a set of marginal events. Additional events can be added if needed but WB2 and DIMOSIC \cite{Magnusson2022, WB2} both used one year of data to address this same challenge. We provide a set of marginal-events for heat waves, extreme freezes, and severe convective days. Details of how these days are selected are provided in the "Methods" section.

\begin{figure}
    \centering
    \includegraphics[width=\linewidth]{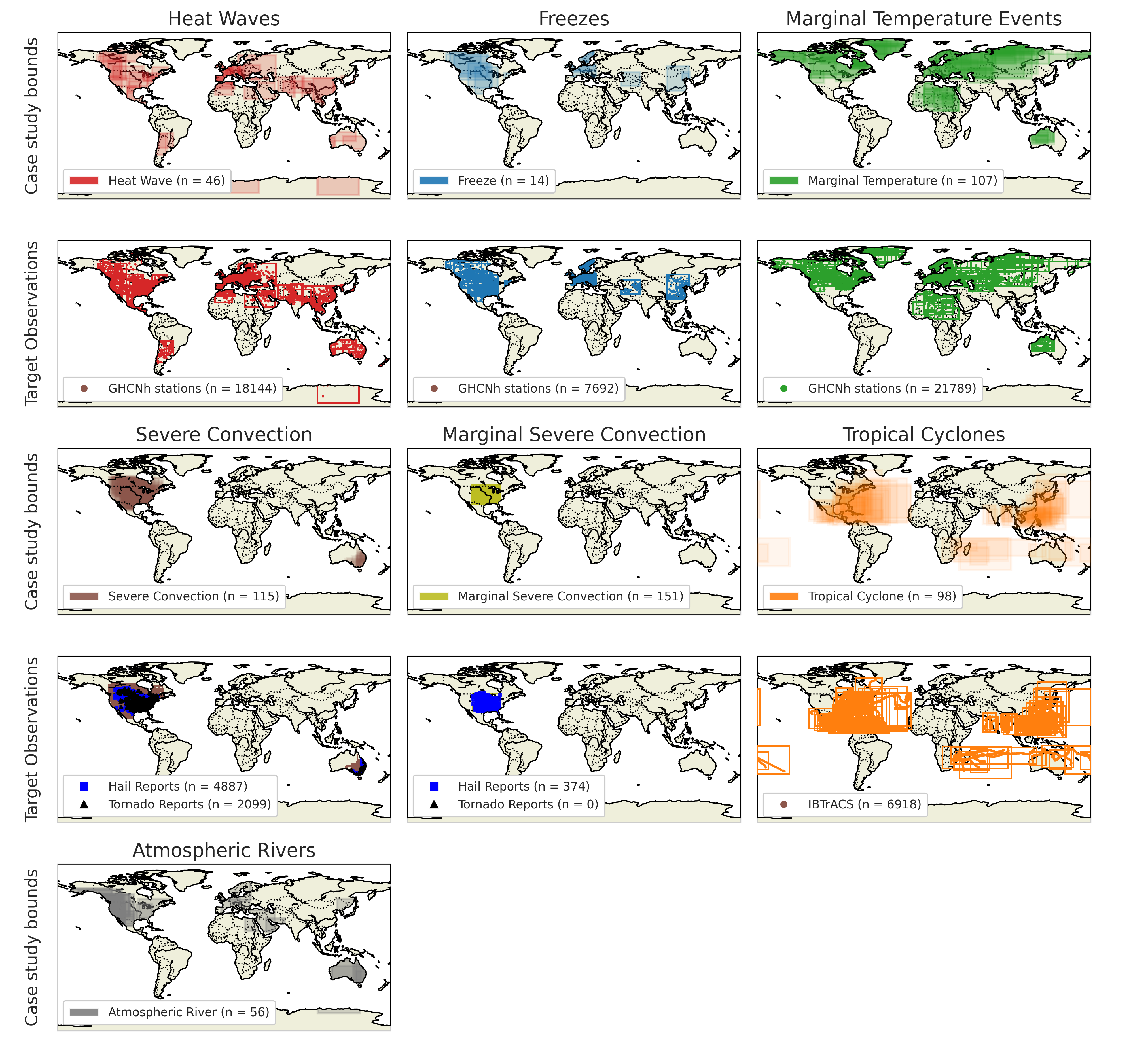}
    \caption{EWB event counts and case regions (left column), target observations (center column), and examples of using the targets (right column).}
    \label{fig:EWB_events}
\end{figure}

EWB contains five main categories of high-impact weather phenomena: heat waves, major large-scale freeze events, severe convective days, tropical cyclones (TCs), and atmospheric rivers (ARs). These categories are drawn from a combination of the NOAA/NCEI study of high-impact events\footnote{\url{https://www.ncei.noaa.gov/access/billions/events} and \url{https://www.ncdc.noaa.gov/stormevents/}}, the WMO\footnote{\url{https://wmo.int/site/world-weather-and-climate-extremes-archive}}, and the database and peril and hazard classification system developed by the Integrated Research on Disaster Risk (IRDR)\footnote{\url{https://www.emdat.be/}}\cite{HazardsIRDR2014}. Figure \ref{fig:EWB_events} shows the locations and regions for each event. Embedded in the map are the numbers of cases for both US and global events. We chose cases from the years 2020-2024 because these are often not used in training the AI global models, thus providing an independent test set.\footnote{While recent AI models sometimes train on the complete ERA5 archive, we argue that holding out 5 years for a consistent testing set is worth the effort to provide a common test set for model comparison.} For each category, we chose case studies to represent the full span of events within the time period. This enables EWB users to evaluate their models for consistent performance and for more statistically robust model comparisons This goal is not always achieved, as some events do not occur frequently across the five year time-period (e.g., freeze events). Figure \ref{fig:EWB_events} also shows two categories of marginal events used to reduce the forecaster's dilemma in practice.

The specific case studies for each category are derived from several sources. For events in the United States, we used the Storm Events database along with the Billion dollar weather disasters\footnote{\url{https://www.climatecentral.org/climate-services/billion-dollar-disasters}} to identify high-impact storms in each category. We also used lists of events available publicly, including on Wikipedia. Comprehensive freely available archives of high-impact weather events outside the United States do not exist. We used a combination of events identified by the ECMWF in their Severe Event Catalogue\footnote{\url{https://confluence.ecmwf.int/display/FCST/Severe+Event+Catalogue}}, events identified by the World Meteorological Organization, events we could find in the news, high-impact events studied in the literature, and events identified by our international collaborators. While this does not lead to a fully comprehensive list, it is sufficient to provide a useful evaluative baseline and we are confident that the list will grow with community involvement in the project.

\subsection{Observations-based verification}\label{observations-based-verification}

EWB takes a unique approach to verification by providing ground-truth observations to the extent possible. Existing approaches primarily compare predictions to a global reanalysis dataset (most frequently ERA5 \cite{ERA5}), due to a lack of publicly available observations. One exception is WeatherReal, a benchmark which utilizes a snapshot of global point observations that have been quality controlled \cite{WeatherReal}. Although ERA5 is considered to be a gold-standard for a global dataset, it contains known biases and non-stationarity through time \cite[e.g.]{ERA5, Lavers2022, Buschow2024} which can be addressed through the use of ground-based observations. EWB prioritizes point observations and data from sources that are not derived from a statistical or physical model (e.g. a reanalysis product), with ERA5 as the fallback option for events or regions without publicly available observations. Ground-based observations are a critical component of understanding the true state of the atmosphere but are also subject to various quality issues, including sensor malfunctions, improper placement, intermittent data outages, and varying correction algorithms depending on the owner \cite{WMOObservations}

EWB incorporates three sets of observations into the potential "targets" (naming convention from WeatherBench). These are shown in Figure \ref{fig:EWB_events} under each of the event categories for heat waves, freeze events, severe convective days, and tropical cyclones. The Global Historical Climatology Network (GHCNh \cite{GHCNh}) network provides hourly surface observations at stations scattered across the globe. EWB specifically makes use of hourly temperature observations to evaluate the heat and freeze cases. Future extreme events could potentially use additional variables including wind and humidity. If there are not sufficient GHCNh stations for a case, the case is not evaluated using GHCNh, although it can be evaluated using ERA5.

The National Oceanic and Atmospheric Administration (NOAA)'s Storm Prediction Center collects a publicly available dataset of all severe storm reports. These results are quality controlled and made available for download. We have also obtained reports for Canada through a collaboration with the Canadian Severe Storms Laboratory, the Northern Hail Project \cite{NHP}, and the Northern Tornadoes Project \cite{NTP}. This unique combined dataset allows us to examine severe convective days across North America. In addition, we have obtained tornado and hail reports for Australia, allowing EWB users to uniquely verify AI model prediction of major severe weather outbreaks across North America and Australia. Additional locations will be added through future collaborations.

The final target dataset is drawn from the International Best Track Archive for Climate Stewardship (IBTrACS, \cite{IBTrACS}). EWB uses the IBTrACS data by default to provide the final cyclone tracks for each of the tropical cyclones in the dataset. These are used to provide verification of track predictions as well as landfall. IBTrACS track data is provided at a temporal resolution of three hours. Genesis and landfall are included as supplementary points in the dataset for each tropical cyclone. Due to the temporal resolution of most global NWP and AIWP models, landfall for both the model tropical cyclone tracks as well as the IBTrACS data is calculated in EWB by taking a linear interpolation between the track points pre-landfall and post-landfall.

Atmospheric rivers are evaluated against ERA5 by computing integrated vapor transport (IVT) for both the forecasts and ERA5 using the specific humidity along with U- and V-component winds from the surface up to 200 hPa \cite{Mo2024}. There are two main reasons we chose to use ERA5 and not bring in additional target data for atmospheric rivers. Most importantly, the United States is the only country to provide large-scale publicly available radar data that has been quality controlled and has nearly complete spatial coverage and very little missing data; when combined with surface observations this produces a reliable climatological precipitation data product. Global measurements of precipitation are satellite derived and not available on the same time frame as current global model forecasts. Second, the majority of global AI weather models are trained using precipitation data from ERA5, which does not directly assimilate precipitation measurements such as those from radar, and often shows large differences when compared to rain gauges \cite{WB2}. As global models evolve to better predict accumulated precipitation, future versions of EWB can include the evaluation of atmospheric rivers using precipitation datasets such as IMERG/GPM.

\subsection{Impact-based Metrics}\label{impact-based-metrics}

EWB provides a standard set of impact-based metrics that evaluate both the \emph{quality} and the \emph{value} of the forecasts. These metrics are chosen with community input to ensure that they both meet the needs of the end-users of the forecasts and are in widespread use worldwide in evaluating forecast quality and value. Additionally, we derive metrics that can generally be computed from any frontier AI weather model output without significant additional post-processing or secondary modeling (such as employing machine learning to generate calibrated severe weather probabilities). Ensuring the forecasts are high \emph{quality} also requires objective evaluation across a large test set with metrics that provide an overall understanding of how well the forecast system is performing broadly. To ensure that the evaluation is meaningful to the end-users, the metrics used for the objective evaluation are chosen to be standard metrics used by forecasters and government agencies when evaluating forecasts for that type of hazard. EWB provides metrics for each category and specific event, along with references to the use of these metrics by worldwide meteorological agencies.

Given the diversity of weather hazards, there is no single metric that will work across all types of weather hazards. While many of the current global AI models use RMSE as a global metric, it does not provide insight into how well a model is performing for most high-impact weather hazards. Even for events where RMSE might make sense, such as temperature-related events, it obscures the impact of a specific event by measuring RMSE across the full globe. RMSE also suffers from the double-penalty effect (i.e., a missing event at validation time \emph{t} compounded with a false event at time \emph{t'}) for temporal errors compared to non-prediction of an event \cite{GillelandRoux2015}. EWB provides a more nuanced approach to evaluating across a wide variety of hazards by choosing metrics targeted to each category of weather phenomena that we provide.

Assessing the \emph{quality} of the forecast is not always the same as ensuring that it has \emph{value} \cite{Murphy1993}. Value comes from the usefulness of the forecast. If a forecast has low overall error but no one is using it, it has very low value. EWB measures value by providing impact-based metrics for each event category. While some impact-based metrics may be used across multiple events, their use is tailored to the event being predicted. The impact-based metrics we provide are derived from the literature and citations are provided for each category and event. We chose each metric to answer specific questions about a forecast and its impact. For example, when evaluating a heat wave, it is important to know the error in both the daily minimum and maximum temperatures, since having a very high minimum temperature for the day means people cannot cool down and can lead to more deaths \cite{AndersonBell2009}.

Any choice of metric will require tradeoffs among competing approaches of how to best evaluate a forecast. EWB provides a set of impact-based metrics for specific choices. However, as an open-source community project, new users can easily extend the set of metrics. For example, for the initial release of EWB, we have focused on deterministic forecast metrics. With community input, we can expand to probabilistic metrics in future work. High-impact events could also be examined from an Eulerian perspective or taking a Lagrangian (flow-following) approach. An Eulerian perspective looks at the impact of the events at the location of the event while a Lagrangian approach would track the event object as it moved (e.g. tracking a tropical cyclone as the forecast shifted). Given our focus on high-impact weather and the value of the forecasts to end-users, EWB currently focuses on the Eulerian perspective but we would welcome additional input on Lagrangian metrics from community collaborators. For example, when predicting a tropical cyclone (TC), it is important to not only predict the strength of the TC correctly but also to predict the speed at which it strengthens, the time and place it will impact people by hitting land, and the associated rainfall and other severe impacts that it will generate.

Table \ref{tab:metrics} summarizes the driving questions for each event and the metrics chosen to answer these questions. Our goal is for EWB to create a common way for people to evaluate AI and NWP models in a way that enables them to understand how they truly compare. This goal also helps to drive the questions we choose. While we list the driving questions and metric names below, the equations for each are given in the Methods section.

\begin{longtblr}[
    caption = {The driving questions overall and for each category of event in EWB and the metrics that answer these questions. Equations for each metric are given in the Methods section.},
    label = {tab:metrics},
]{
    colspec = {Q[wd=0.8in, valign=m] Q[wd=2.1in, valign=m] Q[wd=2.1in, valign=m]},
    hlines,
    vlines,
    row{1} = {font=\bfseries},
    cells = {bg=white},
}
Event & Driving Question & Metrics \\
    \SetCell[r=4]{m} Heat Waves and Cold Extremes
        & How far in advance can the model predict a major heat wave or cold spell?
        & Lead time of heat/freeze event \\
        & What is the aggregate error on the event maximum or minimum temperature?
        & MAE, RMSE, regional RMSE over the event region \\
        & What is the aggregate error on the predicted highest low temperature?
        & MAE, RMSE, regional RMSE over the event region \\
        & How often is the model overpredicting heat/cold?
        & Measured by the same metrics on the marginal event days \\
    \SetCell[r=5]{m} Convective outbreak days
        & How many days in advance does the model predict a convective outbreak day?
        & Lead time for a severe risk falling within the outbreak region \\
        & \SetCell[r=2]{m} How precise is the model in predicting the risk area?
        & Intersection Over Union (IOU)/Critical Success Index (CSI) of the predicted event compared to the PPH region of the actual event \\
        & & False alarm ratio calculated over the area of the predicted event compared to the Practically Perfect Hindcast (PPH) region of the actual event \\
        & What is the spatial precision of the prediction based on actual storm reports?
        & Hits and misses of severe reports \\
        & How often is the model over or under predicting convective risk?
        & Number of days where the model predicts risk but nothing happened, measured on the marginal event days \\
    \SetCell[r=3]{m} Atmospheric Rivers
        & How far in advance can a major AR be predicted?
        & Lead time of AR when the AR first intersects land \\
        & What is the spatial error of the predicted AR on land versus observed AR on land?
        & Spatial displacement of the center of mass of where the AR intersects land \\
        & How well is the predicted area of IVT matched to the area where the AR actually landed?
        & IOU on the predicted versus actual AR objects where they overlap land \\
    \SetCell[r=4]{m} Tropical Cyclones
        & \SetCell[r=2]{m} What is the error on landfall? Both in time and space
        & Spatial error of landfall as a function of lead time \\
        & & Temporal error of landfall \\
        & \SetCell[r=2]{m} How well is the intensity of the TC at landfall predicted?
        & MAE of the surface minimum pressure between TC and predicted TC \\
        & & MAE of the peak wind speed between TC and predicted TC \\
\end{longtblr}

\section{In-depth Case studies}\label{in-depth-case-studies}

EWB enables users to evaluate at different levels of granularity and across spatial and temporal scales. For example, a user can evaluate model performance across all events of a specific category, events in a specific geographic region (e.g. North America), events in a specific time window, or even just focus on one event. By allowing users to evaluate their models across these modalities, EWB enables users to dive deeply into the reasons that an AI model is either predicting well or poorly in a specific region or on a specific high-impact event. This approach will significantly improve understanding of why a model may have gone wrong in specific cases or regimes. We illustrate this on each of the event categories.

We evaluated existing AI models using two archives of AI forecasts. First, we created a multi-year archive of AIFS-Single \cite{AIFS}, Graphcast \cite{GraphCast}, and Panguweather \cite{Pangu} re-forecasts. The details of the archive and the initializations used are in the Methods section. The second archive was the CSU-CIRA archive \cite{Radford2025}, which provided a partial archive of FourCastNet v2 \cite{FourCastNet}\cite{FCNv2SFNO}, Graphcast \cite{GraphCast}, and Panguweather \cite{Pangu}. We also created an archive of ECMWF's Integrated Forecast System (IFS) high resolution forecast (HRES) \cite{IFSHRES} as the reference baseline, as is done in WeatherBench. Building a deeper understanding of how AI models are performing for a variety of extreme weather events can provide model developers, model consumers, and researchers with more information to better understand pathways to betterment and research.

\subsection{Heat Waves}\label{heat-waves}

As the globe is warming, heat waves are becoming more impactful as they are happening at a higher frequency, last longer, and occur in places that traditionally do not experience intense heat \cite{MoreHeatWaves1}, \cite{MoreHeatWaves2}. Heat waves kill people, especially the most vulnerable who often lack the ability to cool down, and also cause disruptions to the electrical grid with increased cooling load \cite{Callahan2026}. A skillful model that can forecast heat many days in advance would be particularly useful to areas that need additional time to prepare for such heat waves. For example, the large-scale heat wave in the summer of 2021 in the Pacific Northwest killed a number of people in an area where homes often lack air conditioning \cite{PNWAirConditioning}. A skillful model will be able to predict the high and low temperatures for the heat wave, as both contribute to the deadliness of these events. When the overnight low does not go below a threshold for humans to cool down, heat waves become even more dangerous.

EWB provides three key metrics for heat waves, as well as the ability to compare the performance of models on non-heatwave events. The ability to compare performance against marginal events is critical to avoiding the forecaster's dilemma where the model overpredicts regularly in order to maximize its score on extreme events.

\begin{figure}
    \centering
    \includegraphics[width=\linewidth]{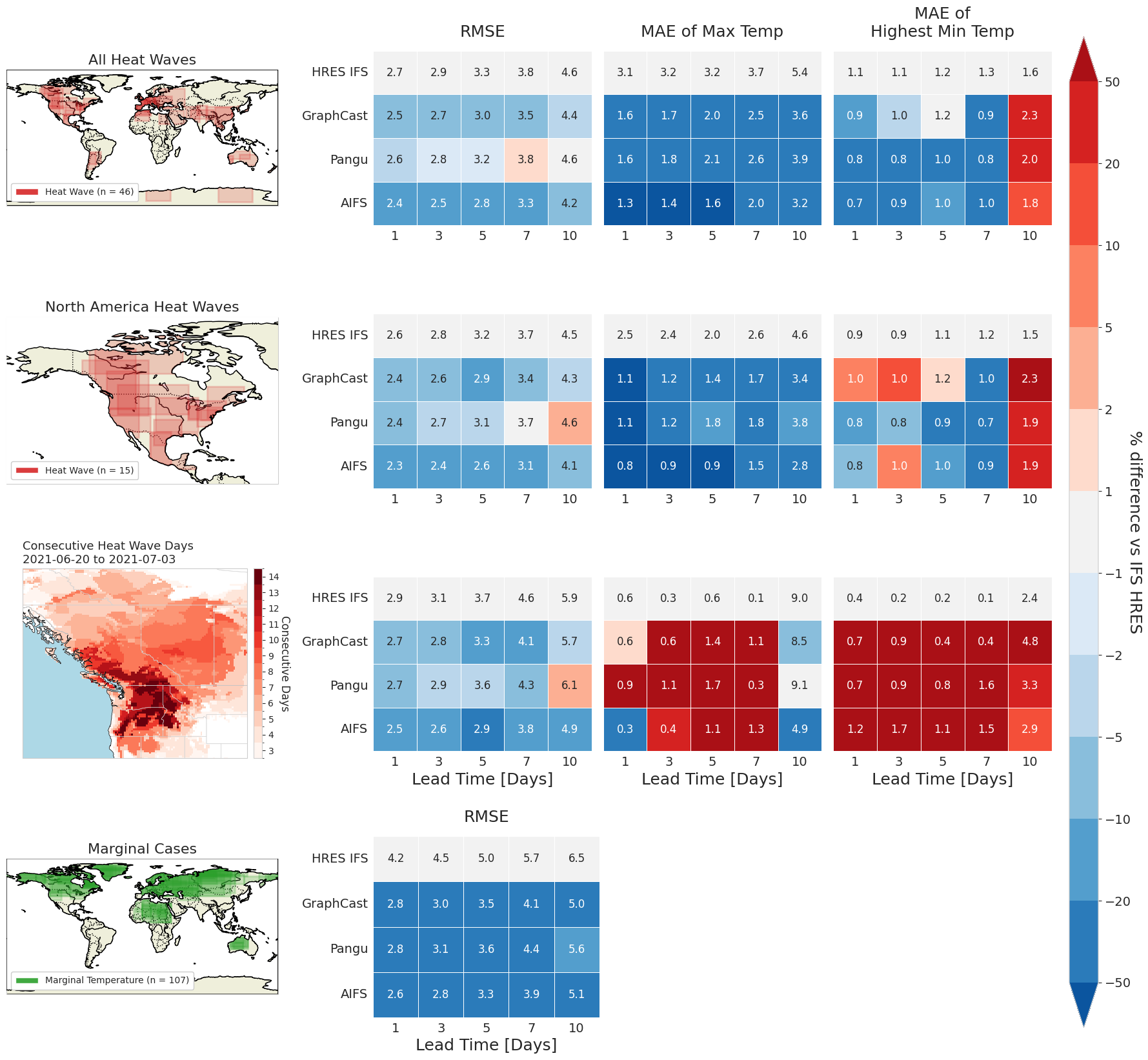}
    \caption{Evaluation of global heat waves (top row), of all heat waves in North America (middle row), and of the Pacific Northwest Heatwave from the summer of 2021 (third row). The bottom row shows the evaluation on marginal temperature events to address the forecaster's dilemma.}
    \label{fig:heat_waves}
\end{figure}

Figure \ref{fig:heat_waves} compares the performance of three current state-of-the-art AI models against the performance of IFS HRES Ensemble Control initialized with IFS conditions. The performance of each model is shown using a heatmap in the style of WeatherBench. While all of the models performed well as measured only by RMSE, their performance varies significantly as a function of region and for the extreme case of the Pacific Northwest. For this case, AIFS was the only model to outperform HRES across all lead times for the predicted event high. None of the AI models were able to outperform HRES for the event low.

Figure \ref{fig:heat_waves} also shows the performance of these models on marginal temperature events, as a comparison point specifically for RMSE. Comparing performance on the marginal temperatures as well as the extremes gives a more complete picture of overall model performance and helps to mitigate a model that only forecasts well for extremes. Overall, all of the models outperformed HRES on the marginal temperature events.

\subsection{Major freeze events}\label{major-freeze-events}

\begin{figure}
    \centering
    \includegraphics[width=\linewidth]{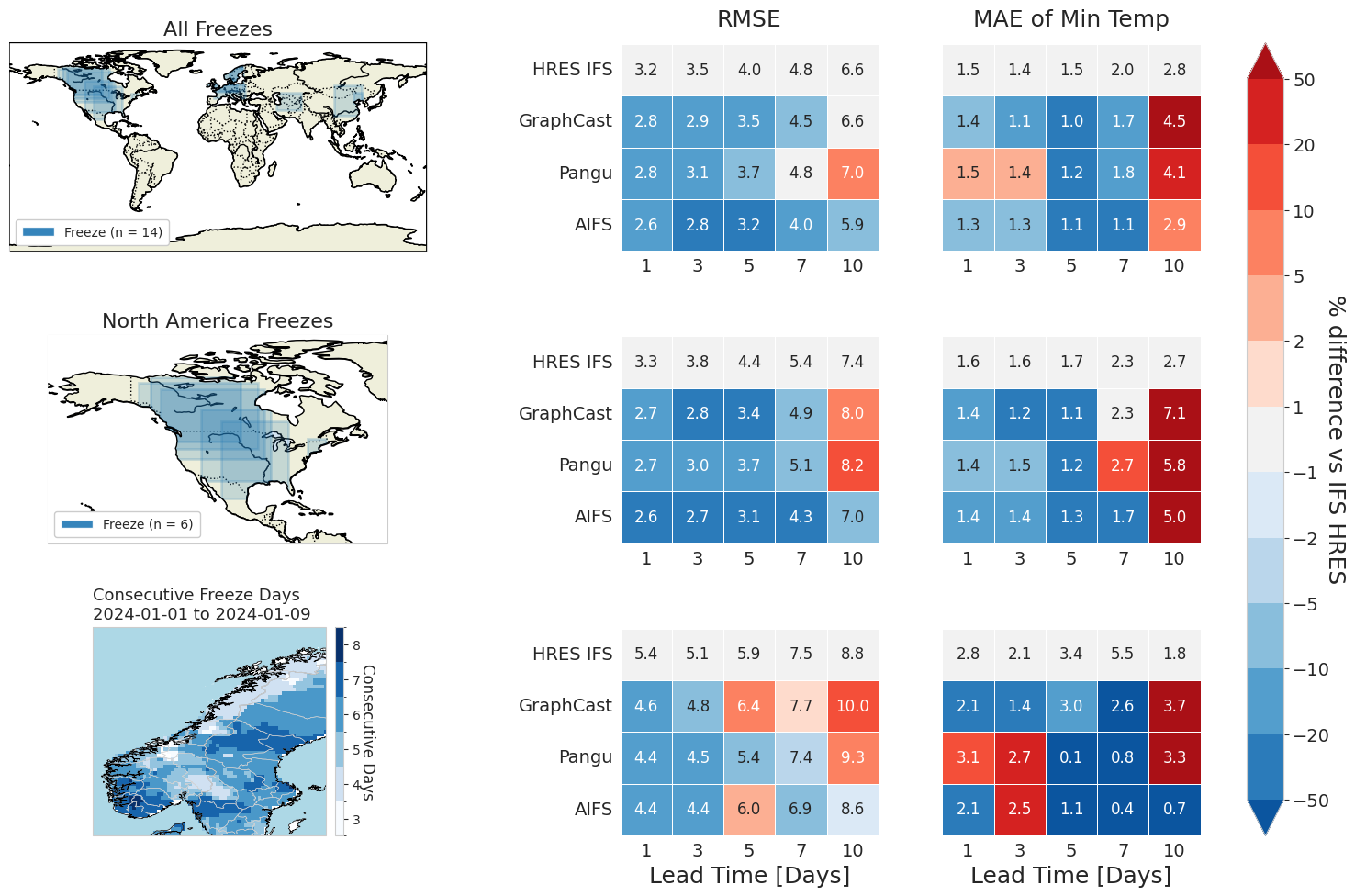}
    \caption{\label{fig:freezes}Evaluation of global freeze events (top row), freeze events in North America (middle row), and a major freeze in Europe in Dec 2022 (bottom row).}
\end{figure}

Similar to major heat waves, major freezes also cause loss of life. In addition, if the freeze events are accompanied by winter precipitation such as freezing rain, there are often widespread disruptions to infrastructure, including electrical outages. EWB evaluates major freeze events using similar metrics to heat waves, examining the predicted low temperature and the RMSE. Figure \ref{fig:freezes} shows the evaluation of the same AI models against EWB's freeze events as well as a set of marginal event winter days. Different from heat events, all of the models struggled to predict the event lows at the longer-time scales, indicating a warm bias or low predictability. AIFS was able to correctly adjust for this warm bias faster both globally and in North America, faster than the other AI models.

\subsection{Convective outbreak days}\label{convective-outbreak-days}

Convective outbreak days provide a unique dataset for AI model evaluation. To our knowledge, although there is one study examining a small number of convective case studies \cite{Feldman2024}, there are no other studies looking at the performance of AI models on large-scale outbreaks. There are recent studies that post-process AI model output to predict convective outbreak days \cite[e.g.]{White2026, Hua2026} but EWB focuses on processing the AI model output directly for the predictions. In addition, EWB provides a unique dataset of storm reports beyond the United States, including both Canadian and Australian storm reports. Together, this enables EWB to examine convective forecast performance at a broader global scale. We acknowledge that there are convective days in additional areas where we were unable to obtain data, but we hope to resolve that with future collaborations.

The metrics for convective days are evaluated in two ways, both of which rely on the storm reports. The first set of metrics directly count the number of storm reports that fall in the area predicted by the model as well as the number of reports that are missed. Because the absence of a report cannot necessarily be counted as there being no severe event, we do not count the other two corners of a standard contingency table (true negatives and false positives). The second set of metrics relies on the Practically Perfect Hindcast \cite{Gensini2020} regions that we create from the reports (see methods for details). Given the coarse spatial/temporal resolution and limited outputs of current AI models, we cannot evaluate individual storms beyond counting if they are in a region or not (previous metric). Instead, we use the area of Craven-Brooks Significant Severe (CBSS) \cite{CBSS} predicted by the models as a proxy for the region where severe convective weather impacts are likely to occur and compare this to the actual PPH area. With these two areas, we can compute the Critical Success Index (also the Intersection over Union of the two areas) and the False Alarm Ratio.

\begin{figure}
    \centering
    \includegraphics[width=\linewidth]{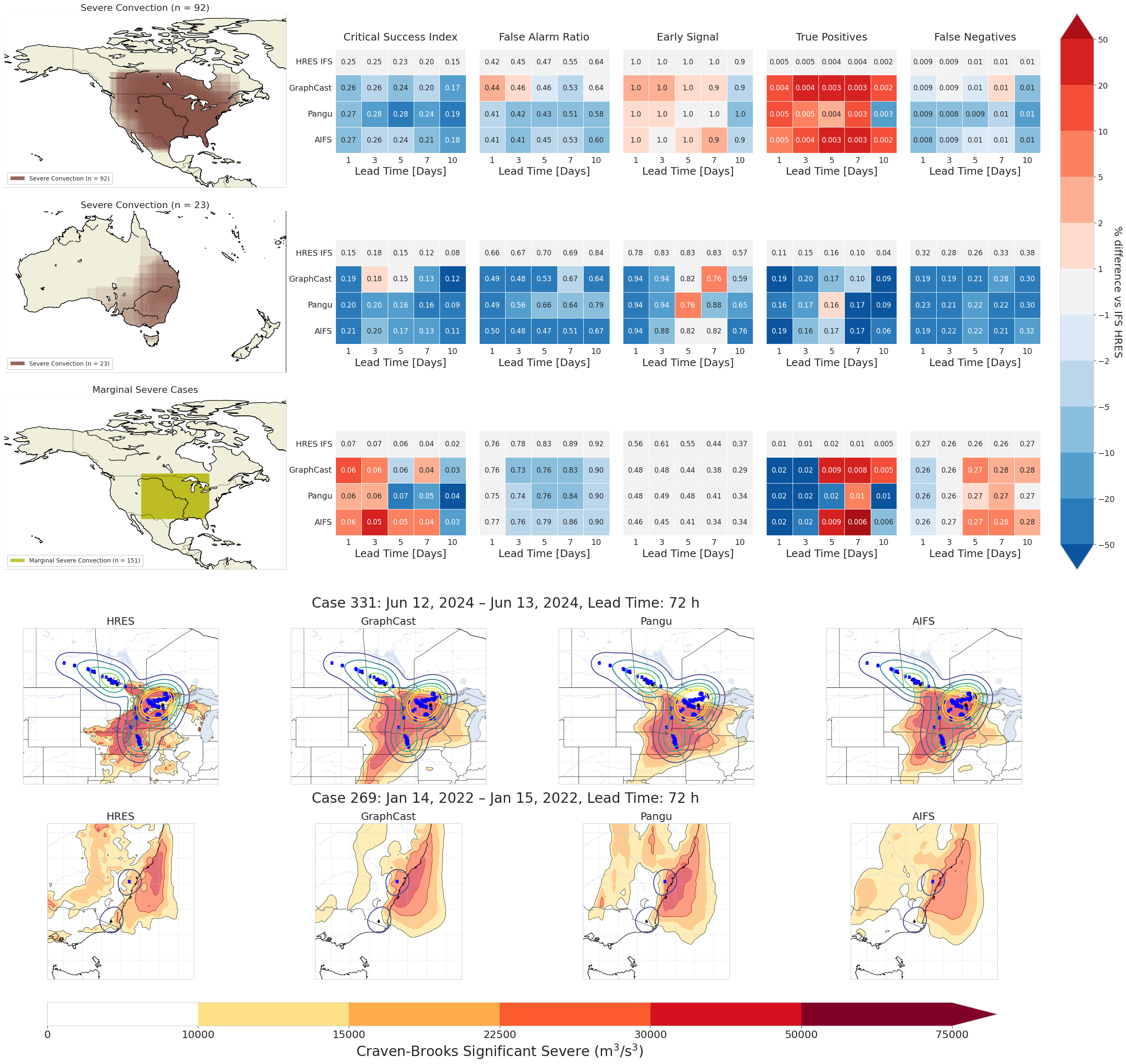}
    \caption{\label{fig:severe_events}Severe convective days in North America (top row), Australia (second row), and marginal severe days (third row). The marginal days early signal coloring is turned off, since it is not clear which direction is ``better'' in detecting marginal events. The bottom two rows show the CBSS 72 hours in advance of an outbreak in North America (fourth row) and an outbreak in Australia (bottom row).}
\end{figure}

Figure \ref{fig:severe_events} shows the results of applying EWB to the severe events globally, across North America and in Australia. Overall, as measured using the PPH areas, the AI models outperform the HRES model for both CSI and FAR. Early Signal detects how early at least fifty percent of the PPH area has a high CBSS predicted for that day. In this case, while most of the AI models do not outperform HRES at the longer lead times, the differences are fairly small. The True Positives and False Negatives are calculated using the severe storm reports for each region. Given the reporting differences between North American and Australia, the true positives and false negatives should not be directly compared between the two regions but rather only compared internally across models. In both cases, the performance differences from HRES is small.

Figure \ref{fig:severe_events} also compares the performance of the AI models on marginal severe days. Here, we want the AI models to predict that there is a risk of severe weather but not as high of a risk as on the days that are outbreaks. Using the same metrics as with outbreaks, we can see that the AI models are overpredicting such days in general.

Finally, Figure \ref{fig:severe_events} plots the CBSS for two severe convective days for each AI model and HRES 72 hours in advance of the event. The PPH contours and LSR points are also shown for each example. For the event in North America (fourth row), while the AI models all predicted a similar area for the event, they overpredicted in the southern part of the region, and both AI and HRES missed the Canadian portion of the event. For the Australian event, only the HRES captured the full southern portion of the event. By providing tools to enable a user to examine convective days both at the more regional scale and at the individual event scale shown here, EWB enables the users to better understand why their model may do well or miss an event.

\subsection{Atmospheric Rivers}\label{atmospheric-rivers}

While atmospheric rivers are an important driver of moisture transport in many parts of the globe, they also bring heavy precipitation, which can cause devastating flooding, snowstorms, and knock-on effects such as avalanches \cite{ARImpacts}. EWB provides both a set of atmospheric rivers for benchmarking and an open-source tracker written in python that can facilitate global tracking and prediction of ARs. As far as we are aware, this feature is unique as the well-known atmospheric tracker tARget is available only in MATLAB \cite{Guan}. Our method utilizes established approaches to tracking ARs \cite{Mo2024}\cite{Newell1992} including utilizing filters for minima of IVT, latitude (to avoid misidentification of TCs as ARs), the Laplacian of IVT, and number of validating gridpoints.

\begin{figure}
    \centering
    \includegraphics[width=\linewidth]{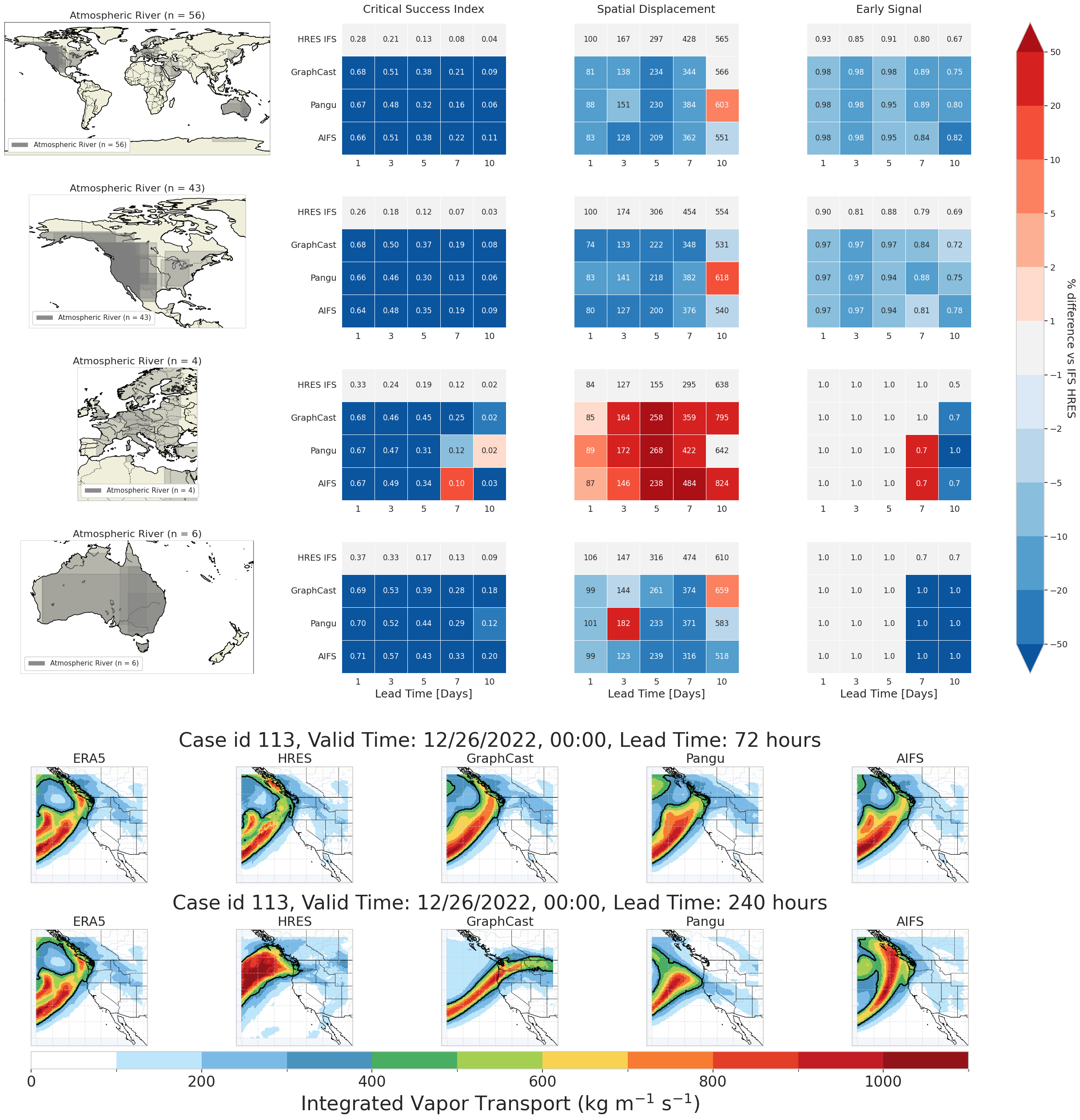}
    \caption{\label{fig:ar_events}Atmospheric river evaluation by region (top 4 rows) and for a specific case study at two lead times (bottom two rows). The top row shows the global evaluation followed by North America, Europe, and Australia.} 
\end{figure}

Figure \ref{fig:ar_events} shows the performance of the AI models as compared to HRES for three key AR metrics. Critical Success Index is computed against the IVT area as compared to ERA5 and for almost all cases, the AI models outperform HRES. Spatial displacement measures the displacement of the center of mass of the AR at landfall. The AI models outperform HRES in North America but struggle in Europe and Australia, although this could be a case of smaller numbers of ARs. Finally, the AI models are able to predict the final area of the AR more frequently at a longer lead time than HRES for North America.

The bottom two rows show the evolution of the prediction for an AR in North America at two different lead times. While all of the models correctly predicted an AR with 10 days lead time, the predicted location of impact as well as the overall shape and intensity vary widely. By 72 hours in advance, the models more closely resemble the final AR as it made landfall. Overall, AIFS represents the AR the closest of the AI-based models but it misses a region of intense IVT that affects British Columbia, whereas HRES was able to correctly predict this regional impact.

\subsection{Tropical Cyclones}\label{tropical-cyclones}

Considering the global impact TCs have on energy transfer from the near-equator to the higher latitudes as well as their impact on human life and property, we created a set of 98 TCs within the period of evaluation of varying intensity. Each TC in this analysis must make landfall due to the evaluation exclusively utilizing landfall-based metrics. One of the unique features of EWB for TCs is that TCs can be evaluated both regionally and individually. Most papers utilizing AI models will evaluate a TC individually or perhaps a small subset of TCs within a basin but comparing across basins and storms is challenging. EWB enables such analysis, which could potentially lead to breakthroughs in TC prediction globally.

\begin{figure}
    \centering
    \includegraphics[width=\linewidth]{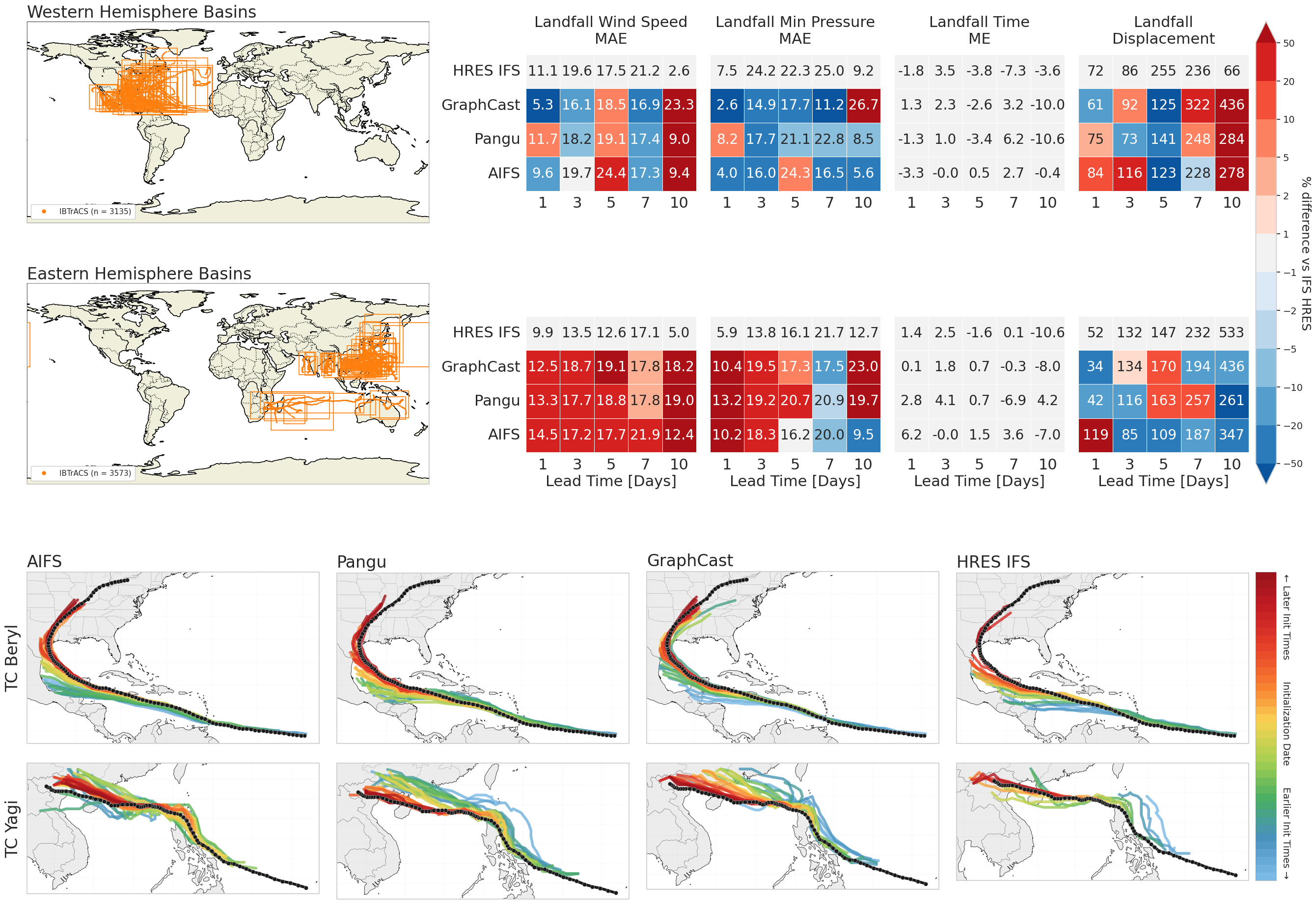}
    \caption{\label{fig:tc_events}Comparison of performance for the Western and Eastern Hemisphere basins for Tropical Cyclones. Bottom two rows compare AI model performance for two storms as a function of model initialization time.}    
\end{figure}

Figure \ref{fig:tc_events} shows the EWB evaluation of the first landfall for tropical cyclones in the Western Hemisphere and the Eastern Hemisphere. Tropical cyclones can have multiple landfalls and EWB can facilitate evaluation of any of these landfalls but we focus on the first landfall, as identified by IBTrACS, for the results shown below. EWB evaluates landfall intensity in two ways: using landfall wind speed and landfall minimum pressure. For the Western Hemisphere TCs, the AI models mostly outperform HRES for the pressure predictions, although performance is worse at longer lead times. Windfall Mean Absolute Error is uniformly worse than HRES in the Eastern Hemisphere basins. Landfall mean error and landfall displacement show mixed results across the hemispheres.

The bottom two rows of Figure fig:TC\_events show the forecast tracks for two tropical cyclones, one in each basin, throughout the lifetime of the TC. The IBTrACS track is shown in black and the coloring of the model tracks shows the initialization time from early to late. For both storms, the earlier initialization times are farther from the actual tracks of the storms but the plots also highlight Graphcast's performance on TC Beryl, where it locked onto the final track fairly early. For TC Yagi, all of the AI models struggled to match the final track early in initialization time.

\section{Discussion and Future work}\label{discussion-and-future-work}

EWB provides a new open-source community-based toolkit to evaluate both physics-based and AI weather prediction models that enables interrogation of performance across a suite of high-impact weather and facilitates inter-model comparison on a standard set of common examples. By providing a common framework, EWB drives the science of model development and prediction forward. To our knowledge, the full suite of events provided by EWB is the first comprehensive, open-source database of curated high-impact weather events that are easily accessible to the research and broader communities. This wide-reaching dataset is intended to provide an objective approach to assess a model\textquotesingle s abilities to predict tails-of-distributions which often reflect the most impactful events to society. We also acknowledge the forecaster's dilemma inherent to focusing on extreme weather events and provide evaluation and listing of ``marginal'' events where possible.

To ensure replicability, all of the events are clearly defined with specific criteria for each. We include open-source Python code and use accessible data (e.g. ARCO ERA5) to enable reproduction of the data and/or to create new events of interest. References are provided with the library to determine initial seed locations and times for all cases. Further, with community involvement, we intend to grow these event listings and expand into new categories, such as droughts, derechos, flooding, blizzards, ice storms, as well as secondary impact events including tornadic outbreaks from TCs, compound flooding, flash droughts, and more. EWB's community and open-source approach provides the opportunity to build new metrics and derived variables which are necessary to analyze such types of events.

Future work includes expanding EWB to evaluate higher resolution models, expanding to evaluation of ensemble models, developing a real-time web interface, and expanding the target observations. As models become higher resolution and limited-area models become more prevalent, it is incumbent for evaluations to be able to scale to accompany these advances. EWB has been built with the ability to scale to different resolution datasets inherent to its pipeline. Probabilistic models are also critical to forecasting. Using proper scoring metrics like the continuous ranked probability score is a natural next step as nascent probabilistic models become more widely available and accessible (given the data challenges present for models with many ensemble members). Analysis is also possible using EWB for seasonal and sub-seasonal forecasting (S2S) as more models are developed to forecast beyond the typical forecast window of medium-range weather models (out to 10-15 days).

\backmatter

\bmhead{Acknowledgements}

This material is based upon work supported by the U.S. National Science Foundation under Grant No. RISE-2019758. Flora was funded by the NOAA/Office of Oceanic and Atmospheric Research under NOAA--University of Oklahoma Cooperative Agreement NA21OAR4320204, U.S. Department of Commerce. Allen was also supported by the U.S. National Science Foundation under Grant No. AGS-1945286. This work comprised regular duties at federally funded NOAA/NSSL for Potvin. The authors thank the Northern Tornado Project and Northern Hail project for sharing Canadian severe storm reports and Iris O'Reilly for her contributions to Extreme Weather Bench. Additional computational and storage resources for developing the datasets and tools for this work were provided by Brightband.

\bmhead{Author Contributions}

A.M. conceived of the initial idea of Extreme Weather Bench and all co-authors helped to refine the idea. T.M. wrote the majority of the code for the benchmarking suite. D.R. developed the MLWP forecast model archive used in this study. A.M. and T.M. created all of the graphics for the paper and the code repository for the paper. A.M. and T.M. also developed the methodologies for case bounds and metadata. N.L. provided substantial support for the metrics, in both implementation and ideas as well as guidance of the statistical justifications such as the forecaster's dilemma. C.P. and M.F. provided guidance on the severe convective outbreak events. J.A. provided the data for severe reports in Australia. A.M. wrote the majority of the paper with input from all of the co-authors.

\bmhead{Data Availability}

The code for Extreme Weather Bench is fully available at \url{https://github.com/brightbandtech/ExtremeWeatherBench}; version 1.0.1 was used to produce the results generated for this manuscript and is available on both GitHub and PyPI. The code for generating all of the plots in this paper is available at \url{https://github.com/amymcgovern/extreme-weather-bench-paper}.

The MLWP simulations performed for this work are publicly available and can be accessed from two sources.

\begin{enumerate}
\def\labelenumi{\arabic{enumi}.}
\item
  A subset of the complete dataset is available via Arraylake on the Earthmover Marketplace \url{https://app.earthmover.io/marketplace/6969541cb7bd57a837d648c8}), hosted via Cloudflare R2. These data are free to access from anywhere around the globe. The subset of data hosted here include the 10m U/V wind speed components, mean sea level pressure, 2m temperature, and the geopotential thickness between 300-500 mb; this subset of data should allow users to replicate the heat wave, freeze, and tropical cyclone event analyses. Arraylake is a managed version of Icechunk, a transactional storage engine built on top of Zarr that provides data curation and management tools as well as improved performance for cloud-native workflows.
\item
  The complete archive of forecasts for each MLWP model is available on Google Cloud Storage from the bucket \texttt{gs://brightband-public-mlwp-forecast-archive} as both a collection of Zarr files and an Icechunk repository virtually referencing these files. Due to egress costs incurred when users download these data to their own compute resources, we have enabled ``requester pays'' on the bucket; users will need an account on Google Cloud Platform to access the data. These fees are lowest (free) if users choose to access the data from compute resources in the us-central1 region on Google Cloud Platform.
\end{enumerate}










\begin{appendices}

\section{Methods}\label{methods}

\subsection{Identifying case studies for each category}\label{identifying-case-studies-for-each-category}

We provide the details of how we selected case studies for each category of event in EWB. There are 5 initial event types included with the library: land-based heat waves, ``cold snaps'' (events causing anomalous freezing conditions for a prolonged period of time), severe convection (e.g. convective mesoscale and synoptic-scale events that produced large hail, tornadoes, and damaging winds), tropical cyclones, and atmospheric rivers. A key part of the definition of each event in EWB is the spatiotemporal shape, or cube, which aligns with both an objective definition of the event type as well as the social perception of the extent of the case.

\subsubsection{\texorpdfstring{Land-based Heat Waves }{Land-based Heat Waves }}\label{land-based-heat-waves}

While we used a combination of reports from the WMO and the news to identify the initial ranges of possible dates for heat waves, we defined the dates and center locations precisely using data from ERA5. To define a heat wave (and the freeze events), we calculated climatologies of surface (2-meter) temperature between 1990 and 2019. Surface temperature using ERA5 data were compared against the 85th percentile of surface temperature, calculated at each gridpoint with a linear decay of 21 days centered on the date of interest, calculated for each hour on a 6 hourly cadence. The dates for the climatology were chosen to be independent of the 2020-2024 evaluation period used for EWB's case studies.

There are many different definitions of a heat wave, but no single objective definition exists\cite{Habeeb2015}. Studies often define a heat wave as prolonged periods exceeding certain percentiles of climatological heat for that location and time. While it is important to include additional variables such as humidity and wind, these are not currently predicted at the surface and thus we focused entirely on temperature for the initial release of EWB. We use the 85th percentile as a baseline to align with the EPA definition of a heatwave \cite{EPAHeatWave}. To be considered a heat wave, a location needs to have at least 3 consecutive days of surface temperature above the 85th percentile. Temporal gaps of no more than 24 hours below the threshold were allowed within the time series for each case study.

\begin{figure}
    \centering
    \includegraphics[width=\linewidth]{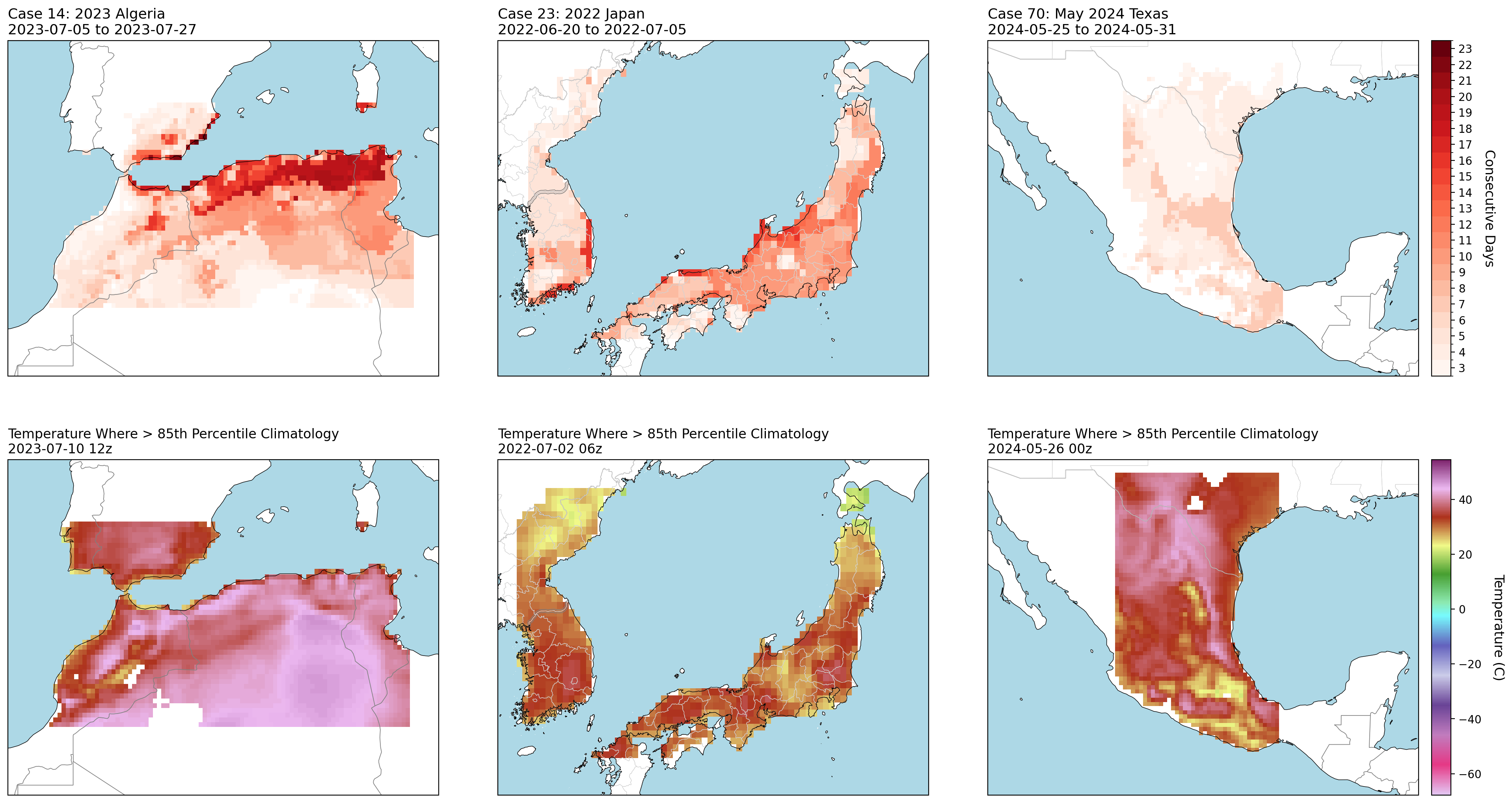}
    \caption{\label{fig:heat_waves_methods}Example of three heat waves and their bounds. The top row shows the cumulative days above the threshold and the bottom row shows the maximum temperature for that heat wave.}    
\end{figure}

We identified the general geographical area for each heat wave based on the WMO and news reports but we chose specific centers for each initially based on the location of interest, such as Seattle for the 2021 Pacific Northwest event. As long as a majority of gridpoints met the criteria, we then iteratively grew each edge of the bounding box by 1 degree and recalculated the consecutive heat wave days until each edge had less than 50\% of the edge grid points with 3 or more consecutive heat wave days. Figure \ref{fig:heat_waves_methods} gives an example for three heat waves.

Due to a lack of target observations and reliable temperature observations over the ocean, ocean grid points are excluded from the error calculations for heat and freeze events.

\subsubsection{\texorpdfstring{Winter weather: Large-scale Major Freezes }{Winter weather: Large-scale Major Freezes }}\label{winter-weather-large-scale-major-freezes}

Large-scale major freezes were defined similarly to the land-based heat waves. Initial locations and time ranges were similarly identified from WMO cases as well as news reports. Criteria uniformly defining an anomalous cold event is significantly more sparse relative to heat waves. We utilize an inverse approach for determining case study bounds compared to heat waves; a day in a major freeze event was required to be both 1) below freezing and 2) below the 15th percentile climatology of the temperature for that gridpoint. As with heat waves, to be considered a valid case study, the temperature for a majority of grid points needed to remain below the 15th percentile for at least 3 days. The physiological impacts of cold snap events are fundamentally different from the coupled apparent temperature impacts of heat waves. Impacts to the built environment, including to the electrical grid, also differ.

\begin{figure}
    \centering
    \includegraphics[width=\linewidth]{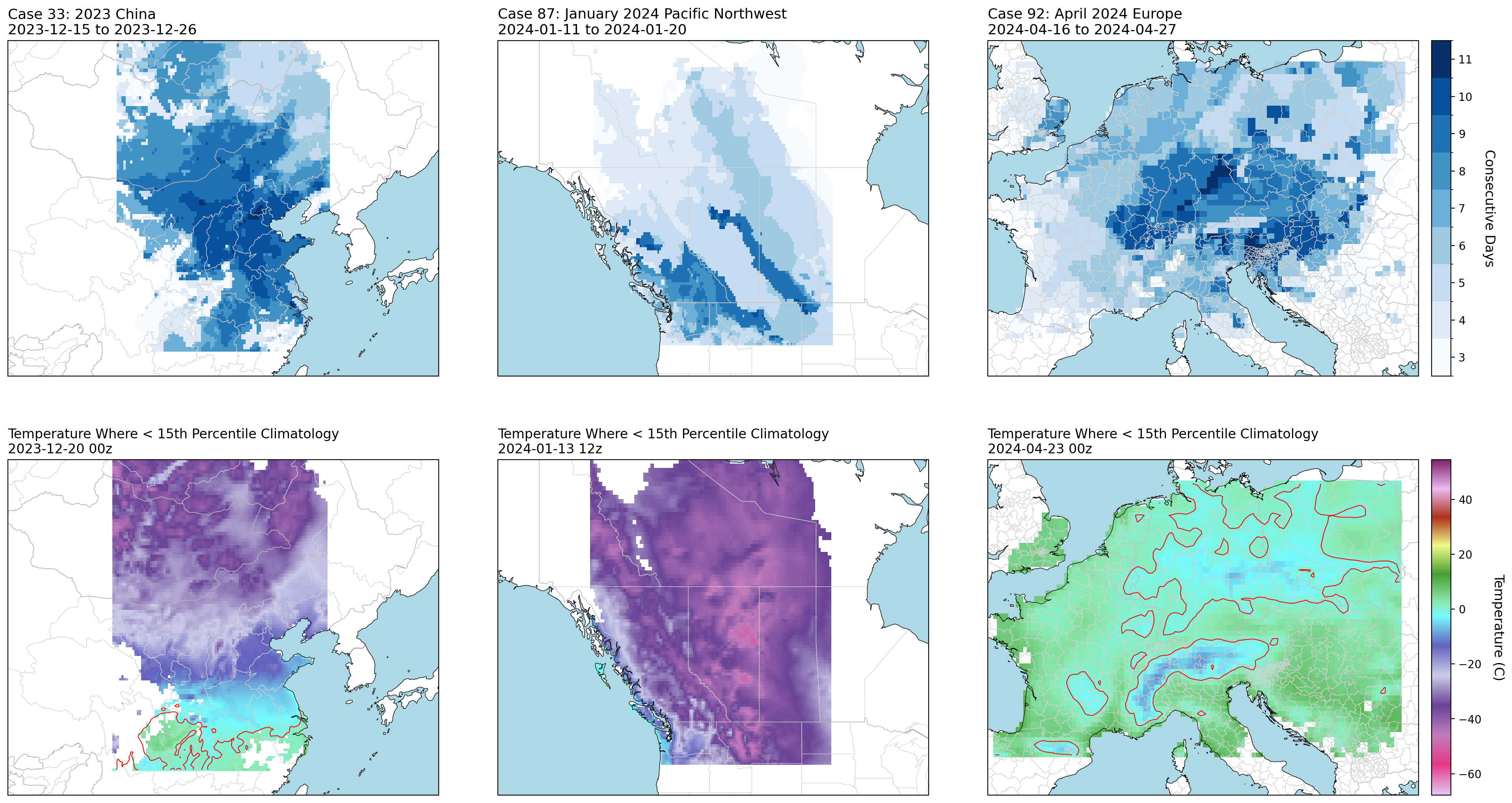}
    \caption{\label{fig:freeze_methods}Example of three freeze events, as in the previous figure.}    
\end{figure}

Figure \ref{fig:freeze_methods} shows an example of three freezes, their minimum temperature and the bounds of each event.

\subsubsection{Land-based marginal temperature days}\label{land-based-marginal-temperature-days}

To address the forecaster's dilemma, we also generate a similar set of marginal events for temperature. Picking random days or periods of several days could potentially overlap existing cases. Instead, we make use of the same idea of climatologies used to define heat and freeze days originally. Marginal temperature days are defined to be regions greater than 200,000 km\textsuperscript{2} where the temperature remains in the 16th to 84th percentile for at least 5 days. We generate these regions automatically from ERA5.

\begin{figure}
    \centering
    \includegraphics[width=\linewidth]{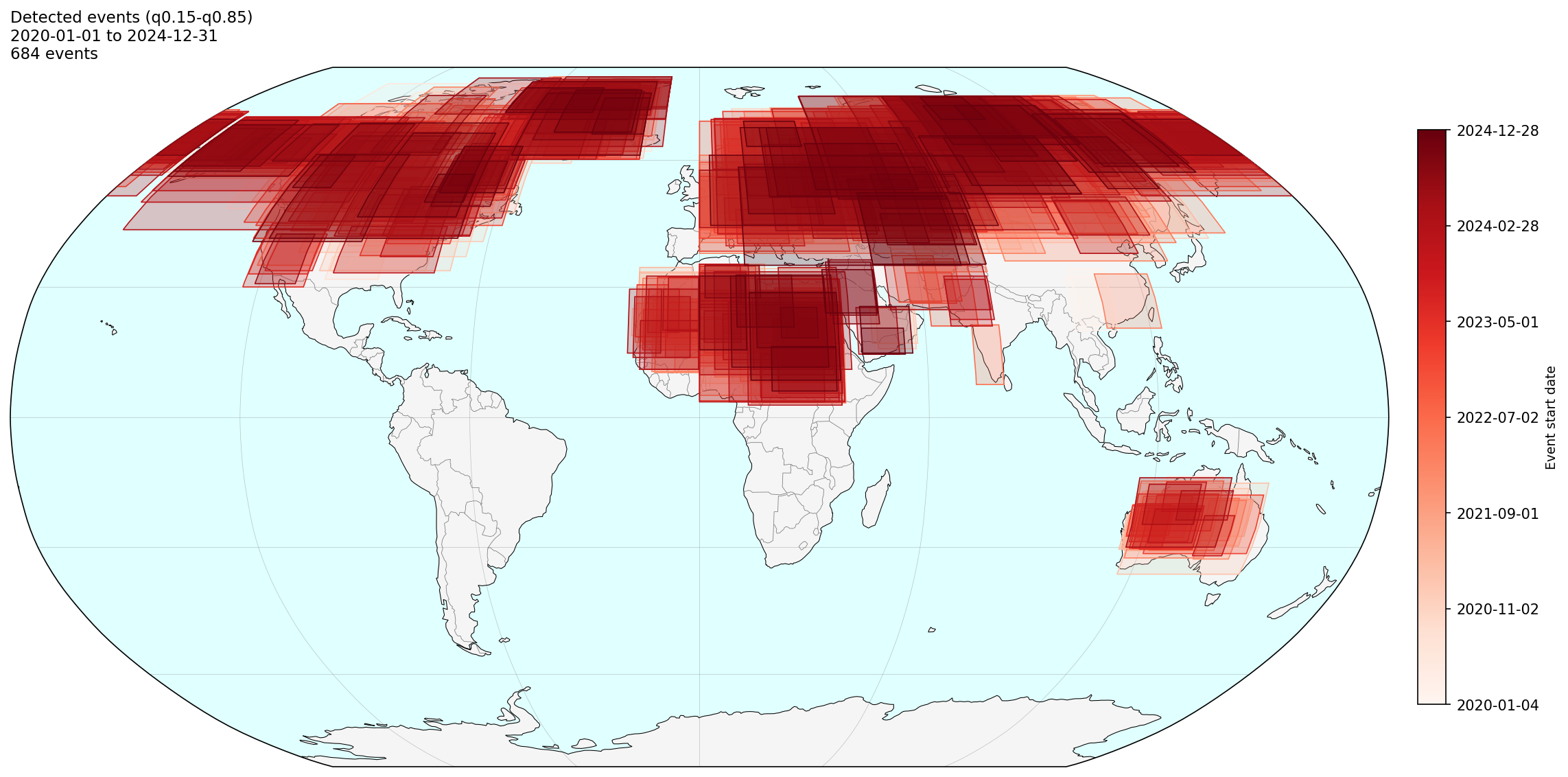}
    \caption{\label{fig:marginal_temp_methods}All marginal days detected by our method.}    
\end{figure}

Figure \ref{fig:marginal_temp_methods} shows all marginal events detected from 2020 through 2024. We excluded Antarctica due to known data biases there. Given the large number of events, we subsampled randomly from this set to create the marginal events used in EWB.

\subsubsection{\texorpdfstring{Convective/Severe Event Days }{Convective/Severe Event Days }}\label{convectivesevere-event-days}

Large-scale convective events are defined inside three countries for the first release of EWB. These countries are the United States, Canada, and Australia. The US has a publicly available dataset cataloging every tornado, hail, and wind report from NOAA's National Centers for Environmental Information (NCEI), called the Storm Events database. To identify the dates in the US for the large-scale convective events, we used the billion dollar disasters database from NCEI as well as the storm reports database from NOAA's Storm Prediction Center (SPC). We selected days with large numbers of tornado, wind, and hail reports. Given that the current resolution of AI models does not facilitate the prediction of individual tornadoes, we felt it was better to focus on days with widespread impact.

Because Canada and Australia reports are more sparse, we applied different criteria to identify severe dates for these areas. For Canada, any date already identified in the US that crossed into Canada was included. If there were Canadian hail or tornado reports, these were included when final bounding boxes were calculated. For days that occurred only in Canada (as computed from the reports from \cite{NTP} and \cite{NHP}), we required there to be at least 10 hail reports and at least one tornado report. If there were severe reports in the US data for this day, they were also included when calculating the bounding box. For Australia, a severe day required a minimum of two reports (hail and tornado combined). While this is a significantly smaller number of reports than the other countries, the Australian data was collected manually by one co-author from news and social media reports that were cross-verifiable with radar data following the procedure described in \cite{Allen2021}, and thus are not as well reported as the US or Canadian databases.

Once severe convective days were identified, we created regions for evaluation by computing Practically Perfect Hindcasts \cite[PPH]{Gensini2020}. Because the density of reports differs across the countries, we created a weighted version of PPH. Whereas the assumption in the US is that nearly all tornados are reported, this is not true for the other databases where population biases can lead to significant underreporting. Thus, we weighted tornado and hail reports at 10 globally.

Once the PPH contours were created, we drew a bounding box around the 0.01 probability contour. Severe convective days can be evaluated in the entire bounding box or only within the PPH region. Most, if not all publicly available global AI weather models as of this paper do not resolve convection explicitly. Determining the best proxy for the model's ability to detect synoptic and mesoscale conditions conducive for severe convective complex formation requires utilizing commonly available model outputs. The Craven-Brooks Significant Severe (CBSS) parameter \cite{CBSS} is a composite index which combines 0-6km bulk shear, calculating using the surface and 500hPa winds with mixed-layer convective available potential energy (MLCAPE). The combination of these two components is a proxy to identify favorable conditions for strong convective growth and was used by \cite{Feldman2024}.

\begin{figure}
    \centering
    \includegraphics[width=\linewidth]{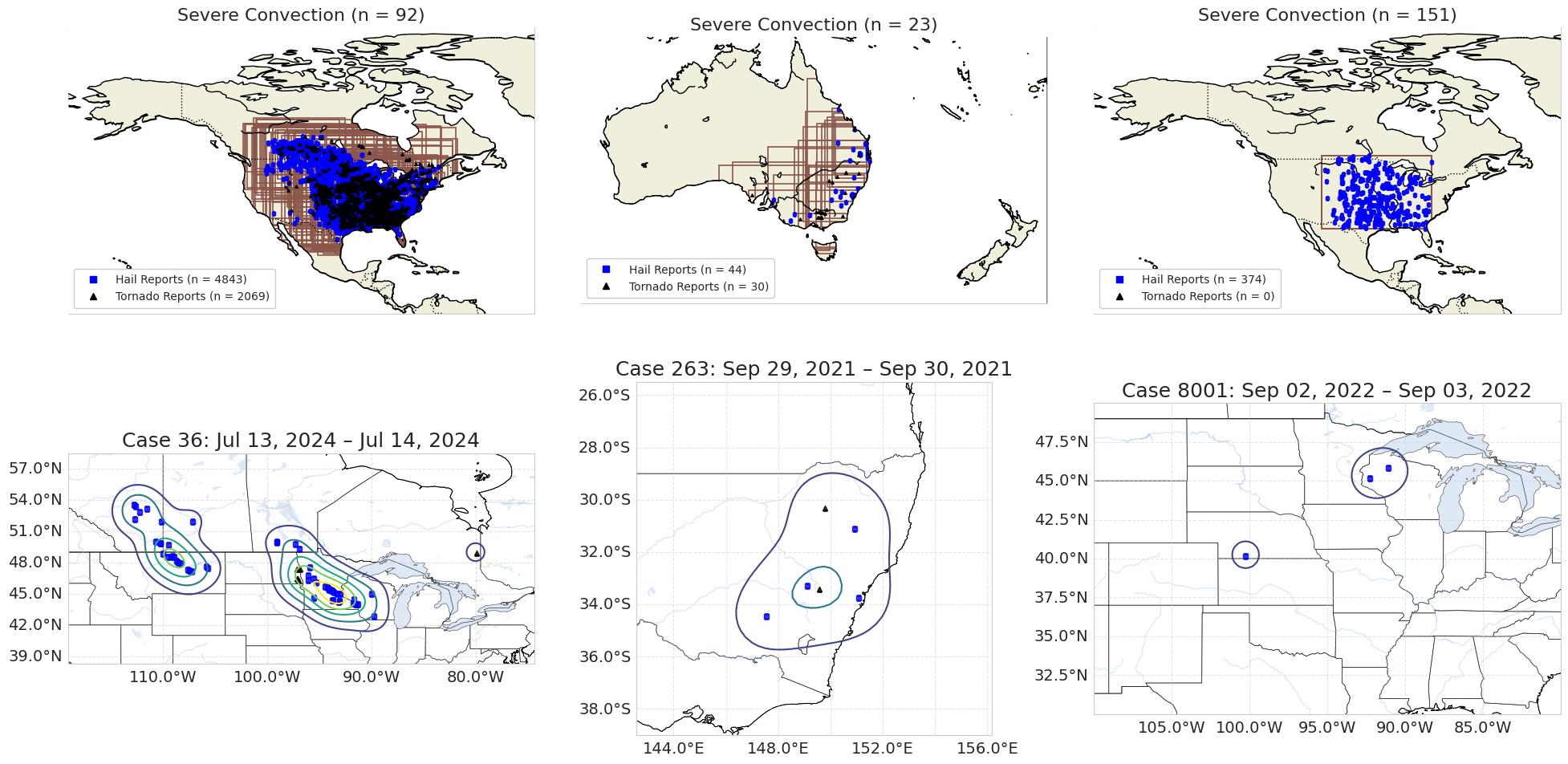}
    \caption{Local storm reports for severe days in North America and Australia (top row, left and middle panel) and marginal severe cases (top right panel). The bottom row shows the PPH contours for selected case studies.}
    \label{fig:severe_methods}
\end{figure}

Figure \ref{fig:severe_methods} shows all of the reports for both North America and Australia (top row) as well as examples in both regions with their reports for that day and the PPH contour that EWB uses for its calculations. This figure shows the stark differences in both the number and density of reports between the two areas. To compute MLCAPE we re-implemented the algorithm from MetPy \cite{May2022} with optimizations for computational efficiency. This implementation is available as part of EWB and can be used as a stand-alone utility.

\subsubsection{Convective/Severe Marginal Event Days}\label{convectivesevere-marginal-event-days}

As with heat and freeze events, we create a comparable marginal event data set for severe convective days. While it could be tempting to choose days with 0 reports, there are two issues with this approach. First, there are reports nearly every day across the US from March through September. Second, such completely clear days do not provide a fair comparison for models against severe convective days. Instead, we chose marginal event days in the US and Canada from days where there are zero tornado reports and less than 10 hail reports. Wind reports are ignored. Given the under-reporting of events in Australia, we did not choose any marginal event days for that region. Bounding boxes for marginal event days are chosen to contain much of the midwestern US and southern border of Canada, where the majority of storms occur. Figure \ref{fig:severe_methods} shows an example marginal day as well as all of the reports across all marginal severe days.

\subsubsection{\texorpdfstring{Atmospheric Rivers }{Atmospheric Rivers }}\label{atmospheric-rivers-1}

Atmospheric River (AR) events were initially identified using the database at the Center for Western Weather and Water Extremes\footnote{\href{https://cw3e.ucsd.edu}{\url{Center for Western Weather and Water Extremes (CW3E)}}} as well as media articles and a handful of publications that focused on specific AR events within our target time period. As with tropical cyclones, ARs were only counted as an event if they made landfall. Similar in diagnoses to heat waves and cold snap events, there is no objective definition for ARs. Different papers identify different tracking approaches \cite[e.g.]{Guan, TempestExtremes, Mo2024} and typically include some form of integrated vapor transport (IVT) maxima, a threshold of minimum number of grid points or area of IVT (e.g. at least 500 gridpoints greater than or equal to 400 kg/m/s), filters to separate other high moisture events such as tropical cyclones, and occasionally aspect ratio filtering to limit the shape profiles of the tracked objects.

\begin{figure}
    \centering
    \includegraphics[width=\linewidth]{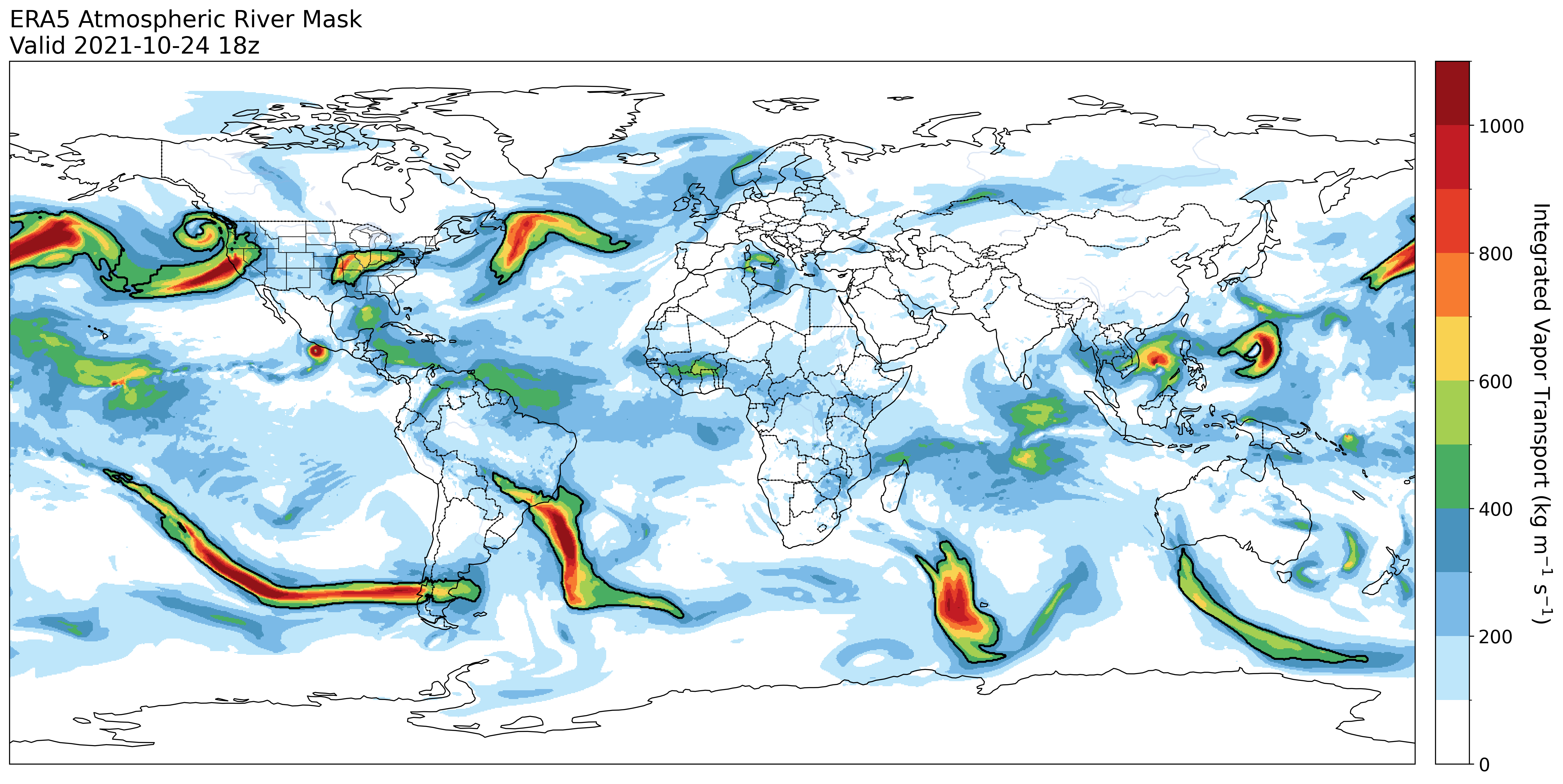}
    \caption{\label{fig:ar_methods}Global atmospheric rivers detected October 24, 2021.}    
\end{figure}

As part of EWB, we developed a native Python-based atmospheric river tracker that includes outputs such as IVT, a boolean mask for valid atmospheric river points, and the intersection of land and an AR mask. The tracker utilizes IVT, the Laplacian of IVT, and a minimum gridpoint amount. We set default values for the code and subsequently this evaluation to be 400 kg/m/s for IVT, the Laplacian of IVT to be 2.5 kg/m\textsuperscript{2}/s\textsuperscript{2} within 8 grid points (for 0.25 degree grid spacing) of the IVT threshold, and at least 500 grid points validating as an AR for each timestep. We use this tracker with the aforementioned defaults to validate selected AR events, build the validation mask for each case of land-intersecting AR (both using ERA5), as well as create a tool that can be used to track global AR events over a predetermined period of time using forecast or reanalysis data. As the AR tracker methodology is based on instantaneous atmospheric quantities, it can be used for any time range dependent on the availability of data and computing resources. Figure \ref{fig:ar_methods} shows a snapshot of the AR tracker globally.

\subsubsection{\texorpdfstring{Tropical Cyclones }{Tropical Cyclones }}\label{tropical-cyclones-1}

Tropical Cyclones (TCs) were identified from online sources that list all named tropical cyclones by basin.We only include land-falling tropical cyclones in our analysis; the total number of TCs is very large so we subsample the full set. Similar to heat and convection, we did not want EWB to only focus on the most intense TCs and thus we included a variety of TC strengths around the globe. We initially included TCs in the mediterranean as one of our basins, however, because Medicanes are so rare, they are not included in the IBTrACS dataset, and thus we had to drop them from EWB.

Tracking TCs has a similar dilemma to ARs in that there's no objective definition on how to track a cyclone's center, although there are plenty of established criteria \cite{WalshTCs}. TCs in global NWP and AI weather models resolve fairly well, but due to the lower resolution compared to mesoscale, convection-allowing, and dedicated TC models, the models are largely unable to resolve the minimum pressure \cite{WalshTCResolution}. Our analysis focuses primarily on other measures of high impact with landfall, including spatial displacement and temporal displacement. Future work on compound hazards from TCs, such as flooding and tornadic outbreaks, are within the scope of EWB.\\
\strut \\
We implement a TC tracker with similar methodologies and variables to TempestExtremes \cite{TempestExtremes} in native Python. The tracker is reliant on analysis data, such as IBTrACS, to restrict the initial spatiotemporal range for genesis as well as subsequent points during tracking. The tracker includes options for various features, including:

\begin{itemize}
\item
  Maximum surface pressure of TC center (defaults to 1020 hPa due to washing out in lower resolutions; practically used to avoid false positive signatures)
\item
  Warm-core feature thresholding using geopotential thickness gradient for 300 to 500 hPa (-6m within 6.5 great circle degrees (GCD))
\item
  Closed contours for surface pressure and geopotential thickness required to be a valid point (True)
\item
  Minimum pressure gradient from identified center (200 Pa within 5.5 GCD)
\item
  Maximum distance from observed track (5 degrees)
\item
  Minimum distance between identified minima at a single timestep (1 degree)
\item
  Maximum time between peaks to be considered a contiguous track (48 hours)
\item
  Total number of timesteps with \textgreater=10 m/s to be considered a valid track (10)
\item
  Highest latitudes to consider a valid point (50N, 50S)
\item
  Distance from the centroid to search for peak winds (2 GCD)
\end{itemize}

These defaults were chosen primarily to maintain similarity to the optimal values proposed within TempestExtremes as well as through sensitivity testing identifying sufficient early stage and dissipating warm-core cyclones, while minimizing spurious ``valid'' candidate cyclone centers.

\begin{figure}
    \centering
    \includegraphics[width=\linewidth]{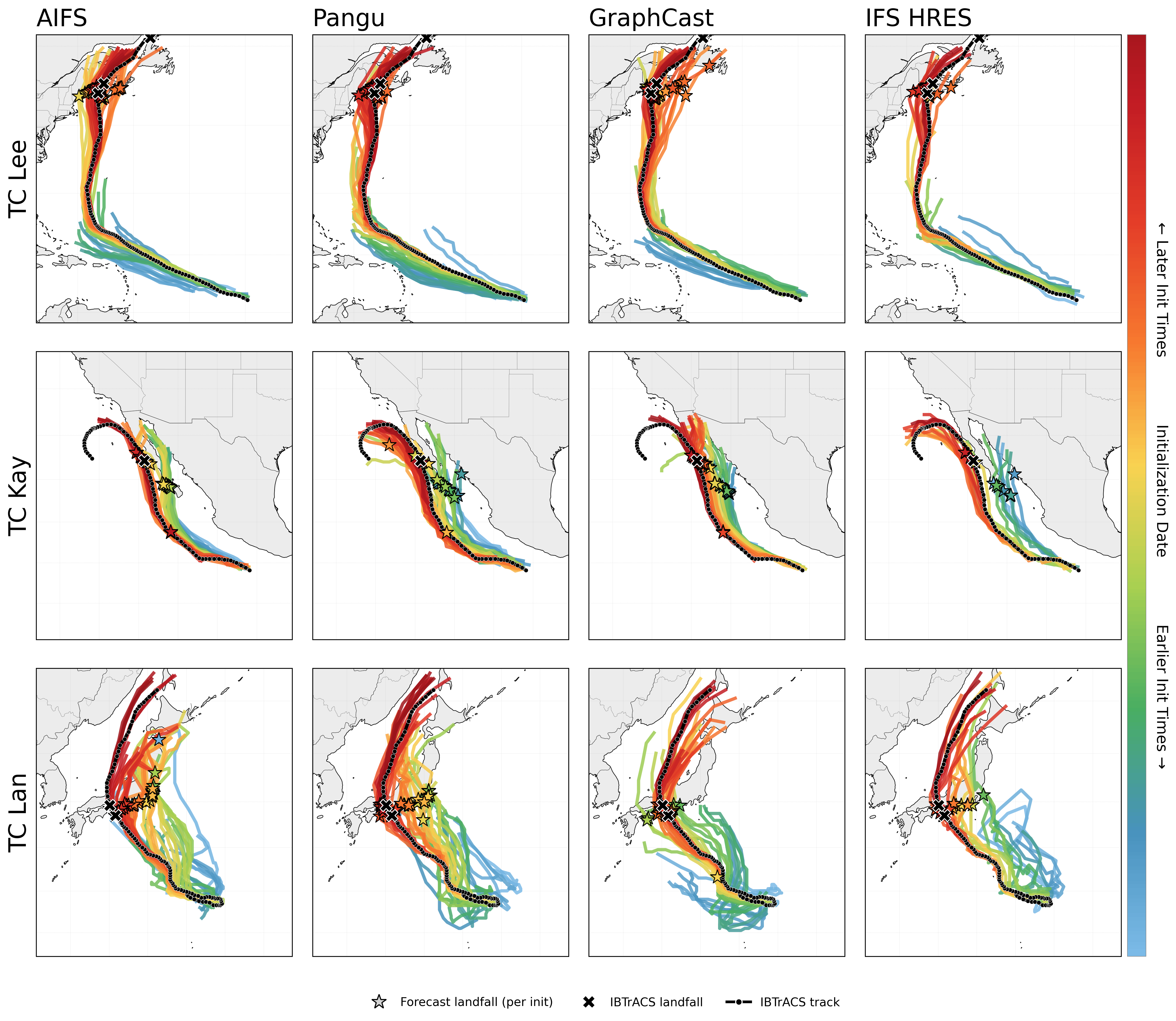}
    \caption{\label{fig:tc_methods}Forecast and IBTrACS analysis tracks for TCs Lee (2023), Kay (2022), and Lan (2023) using the tracking algorithm in EWB for AIFS, PanguWeather, GraphCast, and HRES, respectively. Black circles and x's denote the IBTrACS analysis tracks and landfall locations, respectively. Non-black colors denote the relative initialization time of the model. Shaded stars indicate the first landfall for each track.}    
\end{figure}

Figure \ref{fig:tc_methods} shows an example of the tracking and analysis for three TCs.

\subsection{Metrics}\label{metrics}

All of the metrics implemented in EWB are defined precisely below. We first list the deterministic metrics and then the metrics for probabilistic and ensemble forecasts.

\subsubsection{Lead Time}\label{lead-time}

The lead time of an event can be computed at the current forecast interval (6 hours) or on a daily basis. We currently compute the lead time in days. The lead time is computed as:

Lead = predicted start day of event - actual start day of the event.

\subsubsection{Mean Absolute Error and Related Metrics}\label{mean-absolute-error-and-related-metrics}

The Mean Absolute Error (MAE) is the average of the absolute error of the difference between the forecasted value (f) and the observed value (o). This is defined precisely as

$$\mbox{MAE}  = \ \frac{1}{N}\sum_{i = 1}^{N}\left| f_{i} - o_{i} \right|,$$

where $i$ is an example.

For EWB, we primarily compute Regional MAE (RMAE). This is the MAE computed over all points within the regional boundary defined for each event.

For the heat and cold events, we compute the RMAE of the maximum forecasted temperature,

$$\mbox{RMAE}_{\max} = \frac{1}{N}\sum_{i=1}^{N}\left| \argmax_{f_i} - \argmax_{o_i} \right|,$$

Where the valid time of the maximum observed temperature is not necessarily the same as the forecast, thus a relaxation criteria of +-24 hours is allowed about the observed maximum temperature in case of phase errors in the forecast. This relaxation is permitted as the metric's impact is intended to convey the overall ability for a model to capture the maximum temperature of an event, not its timing. It also helps in avoiding the double-penalty in errors due to an event that validates but at the incorrect time. We also compute the RMAE of the highest minimum temperature forecast for each case,

$$\mbox{RMAE}_{maxdailymin} = \frac{1}{N}\sum_{i=1}^{N}\left| \max_{i \in N}\left(\min_{i \in D}(f_{i})\right) - \max_{i \in N}\left(\min_{i \in D}(o_{i})\right) \right|,$$

Where the maximum of each day \emph{D}'s minimum temperature is calculated. Similarly to RMAE\textsubscript{max}, there is also a +-24 hour relaxation criteria to allow for a phase shift of a day in either direction. It is critical that models capture elements of a heat wave that are most consequential to physiological effects, thus we include a derived metric that determines the error of nighttime peaks of a heat event.

\subsubsection{Intersection Over Union (IOU)}\label{intersection-over-union-iou}

For events with a large spatial extent, such as severe weather or atmospheric rivers, the intersection over union (IOU) metric provides critical information about the overlap between the forecasted areas and the observed event areas. The optimal value for IOU is 1, meaning the predicted event area and observed event area overlap perfectly. If the value is 0, then the predicted area does not overlap the observed area.

$$IOU\  = \ \frac{Area\ of\ \cap_{predicted\ event}^{observed\ event}}{Area\ of\ \cup_{predicted\ event}^{observed\ event}}$$

\subsubsection{Spatial displacement}\label{spatial-displacement}

Events such as tropical cyclones have a specific location for landfall and other events have a more diffuse location, such as severe weather. To make the calculation standard across all event types, the spatial displacement is calculated from the center of the area for the event. The center of the area is calculated by averaging the vertices that comprise the event polygon. While we acknowledge that computing distance using Euclidean distance could lead to biases over larger areas, the biases will be equally applied across all models in EWB.

$$D = \sqrt{(f_{lat} - o_{lat})^2 + (f_{lon} - o_{lon})^2}$$

\subsubsection{Hits and Misses for Severe Weather Reports}\label{hits-and-misses-for-severe-weather-reports}

While it would be ideal to calculate all of the four parts of the standard contingency table, current AI models are not forecasting at a resolution that enables all of these to be computed. Instead, we can use the storm reports and the predicted area of severe weather to compute hits (true positives) and misses (false negatives).

\subsubsection{Practically Perfect Hindcast}\label{practically-perfect-hindcast}

The practically perfect hindcast (PPH) region is defined to be a two-dimensional Gaussian kernel with a defined standard deviation $\sigma$, that is derived from the LSR data \cite{Gensini2020}. For the purposes of this paper, the analysis excluded wind reports as many reports occur downwind of the primary regions of CBSS due to propagating mesoscale convective systems (MCS) and/or quasi-linear convective systems (QLCS, often referred to as derechos), one-off convective cells by means of microbursts, or other straight-line winds.

\subsubsection{\texorpdfstring{False Alarm Ratio for Severe Weather }{False Alarm Ratio for Severe Weather }}\label{false-alarm-ratio-for-severe-weather}

The False Alarm Ratio (FAR) is a measure of the false alarms which occurred relative to the full suite of positive predictions in the contingency table; i.e., on a scale of 0 to 1, 0 is a ``perfect'' forecast by predicting no false alarms. We calculate the false alarm ratio for severe weather days within the region defined by the \(CBSS\  \geq \ 15,000{}^{3}{}^{- 3}\) and a \(PPH\  \geq \ 0.01\) (unitless) \(CBSS\  \geq \ 15,000{}^{3}{}^{- 3}\)..

$$FAR = \frac{FP}{(TP + FP)}$$

\subsubsection{CSI for Severe Weather Regions}\label{csi-for-severe-weather-regions}

Although it is not possible to calculate the true CSI based on specific tornado forecasts, we can approximate it by calculating the CSI based on the regions calculated from the Practically Perfect Hindcast \cite{Gensini2020} and the region predicted by the model from CBSS. Once these two regions are identified, they can be binarized into a yes/no prediction using a threshold. For EWB, we binarize using \(CBSS\  \geq \ 15,000{}^{3}{}^{- 3}\) and\(PPH\  \geq \ 0.3\)01. With the binary areas, we can compute all four values of the standard 2x2 contingency table, thus enabling all contingency table statistics to be calculated.

$$CSI = \frac{TP}{TP + FN  + FP}$$

\subsection{Landfall Calculations for Tropical Cyclones}\label{landfall-calculations-for-tropical-cyclones}

There are two components to the TC evaluation, tracking of the TC itself and the determination of landfall. Finding the landfall involves identifying the intersection of a point in forecast time \emph{t} and \emph{t+1} of a valid TC track where the track moves from ocean to land (or ocean to ocean where there is land at some point in the line segment, e.g. for islands, fast-tracking storms, peninsulas, etc.). The land mask is a 1:50m scale map to avoid detecting landfalls on minute features such as atolls which are typically not considered landfall events.\\
\strut \\
To maintain fairness between global models that produce output at a 6-hourly time resolution, and an analysis dataset that is not subject to time resolution limitations - the landfalling time is recorded precisely in the dataset - we linearly interpolate between the aforementioned \emph{t} and \emph{t+1} of both a model's intersecting line segment, as well as a upscaled IBTrACS (as it is also at a 3-hourly cadence notwithstanding landfall and genesis). The upscaling is performed by simply aligning the dataset to the model output.

Landfalls for forecast models have a significant amount of variability and edge cases. To maintain evaluation on relevant landfalls, we restrict valid landfalls using the following criteria:

\begin{enumerate}
\def\labelenumi{\arabic{enumi}.}
\item
  Each valid track will only select the next landfall; i.e. if there is more than one landfall in the forecast track or target track, the first one is selected to evaluate
\item
  Secondary landfalls within 50km of a previous landfall are removed for both forecast and target data
\item
  Landfalls are dropped which occur between the model's initialization time and the track's starting valid time
\item
  Forecast landfalls must be within 24 hours of target landfalls to previous spurious landfall comparisons in regions of multiple land masses \cite{TC24HourFilter}
\end{enumerate}

Finally, an option is included to allow for either the ``first'' landfall, i.e. evaluate only the landfall which occurs first in the analysis data, or the ``next'' landfall, i.e. evaluate the next consecutive landfall in the analysis data only. These approaches are provided to avoid significant amounts of spurious landfalls. A future analysis can augment this methodology to allow for ``all'' landfalls and assess frequency of errant landfalls. Four metrics are then used to evaluate TCs: MAE of the minimum surface pressure, maximum wind speed (within 2 GCD of the center), ME of landfall time, and spatial displacement of landfall. Spatial displacement is calculated as the haversine distance between the forecast and target landfall points. It is expected that spatial resolution issues will result in non-negligible errors for the central pressure of a TC, thus the approach to compare against IFS HRES is similarly valid instead of emphasizing the magnitude of the error itself.

\subsection{MLWP Modeling}\label{mlwp-modeling}

For this work we created a multi-year archive of AIFS-Single \cite{AIFS}, Graphcast \cite{GraphCast}, and Pangu-Weather \cite{Pangu} re-forecasts. All of our forecasts were initialized from HRES analysis retrieved from the ECMWF Open Data archive on Google Cloud Storage. No modifications were made to the reference model source code, however we did build common post-processing code that adapted model outputs to follow CF-Conventions for field metadata.

\subsubsection{AIFS-Single}\label{aifs-single}

AIFS-Single forecasts utilized version 1.0 of the model weights, which were retrieved from ECMWF's repository on HuggingFace (https://huggingface.co/ecmwf/aifs-single-1.0). The implementation of the scripts used to drive the forecasts was adapted from the demonstration notebook on that repository. These forecasts differ from the AIFS-Single forecasts on the ECMWF Open Data in one significant way: as in the demonstration notebooks, we regridded the HRES analysis from its native 0.25 degree grid to the N320 grid used by the model, whereas the operational forecasts run by ECMWF pull from their higher-resolution native analysis. All model outputs were regridded back to the 0.25 degree grid using linear interpolation.

\subsubsection{GraphCast}\label{graphcast}

GraphCast forecasts were generated using the published weights from Google DeepMind's public cloud storage bucket at \texttt{gs://dm\_graphcast}, specifically the ``GraphCast\_operational'' weights. We adapted the demonstration source code for the model from the Google DeepMind GitHub repository at \href{https://github.com/google-deepmind/graphcast}{https://github.com/google-deepmind/graphcast}.

\subsubsection{Pangu-Weather}\label{pangu-weather}

Pangu-Weather forecasts were generated using the published model weights, normalization metrics, and reference source code documented at the GitHub repository, \href{https://github.com/198808xc/Pangu-Weather}{\url{https://github.com/198808xc/Pangu-Weather}}. All of our forecasts were generated using the ``hierarchical'' roll-out strategy, leveraging combinations of both the 24-hour and 6-hour model weights.

\end{appendices}


\bibliography{refs}

\end{document}